\documentclass{article} % For LaTeX2e
\usepackage{iclr2026_conference,times}

\usepackage{preamble}

\title{Latent Stochastic Interpolants}

\author{%
  Saurabh Singh \thanks{Work done while at DeepMind.}\\
  Poetiq AI\\
  \texttt{saurabh@poetiq.ai}\\
   \And
   Dmitry Lagun \\
   Google DeepMind \\
   \texttt{dlagun@google.com} \\
}
\iclrfinalcopy % Uncomment for camera-ready version, but NOT for submission.
\begin{document}

\maketitle

\begin{abstract}
Stochastic Interpolants (SI) is a powerful framework for generative modeling,
capable of flexibly transforming between two probability distributions.
However, its use in jointly optimized latent variable models remains unexplored
as it requires direct access to the samples from the two distributions.
This work presents Latent Stochastic Interpolants (LSI) enabling joint learning
in a latent space with end-to-end optimized encoder, decoder and latent SI
models. We achieve this by developing a principled Evidence Lower Bound (ELBO)
objective derived directly in continuous time.
The joint optimization allows LSI to learn effective latent representations
along with a generative process that transforms an arbitrary prior distribution
into the encoder-defined aggregated posterior.
LSI sidesteps the simple priors of the normal diffusion models and mitigates the
computational demands of applying SI directly in high-dimensional observation
spaces, while preserving the generative flexibility of the SI framework.
We demonstrate the efficacy of LSI through comprehensive experiments on the
standard large scale ImageNet generation benchmark.
\end{abstract}

%%%%%%%%%%%%%%%%%%%%%%%%%%%%%%%%%%%%%%%%%%%%%%%%%%%%%%%%%%%%%%%%%%%%%
\section{Introduction}

Diffusion models have achieved remarkable success in modeling complex, high-dimensional data distributions across various domains. These models learn to transform a simple ``prior'' distribution $p_0$, such as a standard Gaussian, into a complex data distribution $p_1$.
While early formulations were constrained to use specific prior distributions that are L\'evy Stable, recent advancements, particularly Stochastic Interpolants (SI) \citep{albergo2023stochastic} offer a powerful, unifying framework capable of bridging arbitrary probability distributions.
However, SI assumes that both the prior $p_0$ and the target $p_1$ distributions are fixed and the samples from both are directly \emph{observed}.
This requirement limits their use in jointly learned latent variable models where the generative model is learned, along with an encoder and a decoder, in a latent unobserved space.
Further, the latent space, often lower dimensional, evolves as the encoder and decoder are jointly optimized.
Lack of support for joint optimization implies that arbitrary fixed latent representations may not be optimally aligned with the generative process resulting in inefficiencies.

To address this, we present Latent Stochastic Interpolants (LSI), a novel framework for end-to-end learning of a generative model in an \emph{unobserved} latent space.
Our key innovation lies in deriving a principled, flexible and scalable training objective as an Evidence Lower Bound (ELBO) directly in continuous time.
This objective, like SI, provides data log-likelihood control, while enabling scalable end-to-end training of the three components: an encoder mapping high-dimensional observations to a latent space, a decoder reconstructing observations from latent representations, and a latent SI model operating entirely within the learned latent space.
Our approach allows transforming arbitrary prior distributions into the encoder-defined aggregated posterior, simultaneously aligning data representations with a high-fidelity generative process using that representation.

LSI's single ELBO objective provides a unified, scalable framework that avoids the need for simple priors of the normal diffusion models, mitigates the computational demands of applying SI directly in high-dimensional observation spaces and offers an alternative to ad-hoc multi-stage training.
Our formulation admits simulation-free training analogous to observation-space diffusion and SI models, while preserving the flexibility of SI framework.
We empirically validate LSI's strengths through comprehensive experiments on the challenging ImageNet generation benchmark, demonstrating competitive generative performance and highlighting its advantages in efficiency.

Our key contributions are: 1) \textbf{Latent stochastic interpolants (LSI):} a novel and flexible framework for scalable training of a latent variable generative model with continuous time dynamic latent variables, where the encoder, decoder and latent generative model are jointly trained, 2) \textbf{Unifying perspective:} a novel perspective on integrating flexible continuous-time formulation of SI within latent variable models, leveraging insights from continuous time stochastic processes, 3) \textbf{Principled ELBO objective:} a new ELBO as a principled training objective that retains strengths of SI -- simple simulation free training and flexible prior choice -- while enabling the benefits of joint training in a latent space.

%%%%%%%%%%%%%%%%%%%%%%%%%%%%%%%%%%%%%%%%%%%%%%%%%%%%%%%%%%%%%%%%%%%%%
\section{Background}
\label{sec:background}

\paragraph{Notation.}
We use small letters $x, y, t$ etc. to represent scalar and vector variables,
$f, g$ etc. to represent functions, Greek letters $\beta, \theta$ etc. to
represent (hyper-)parameters. Lower case letters $x$ are used to represent both
the random variable and a particular value $x \sim p(x)$. Dependence on an
argument $t$ is indicated as a subscript $u_t$ or argument $u(t)$ interchangeably.

Our work builds upon two key results briefly reviewed below. The first result
\citep{li2020scalable,theodorou2015nonlinear} states an Evidence Lower Bound (ELBO)
for models using continuous time dynamic latent variables.
We state a more general form than the original to aid the discussion of
the prior distributions.
The second result is a well known method for constructing a stochastic mapping
between two distributions. We exploit it to construct a variational approximation
in the latent space.

\subsection{Variational Lower Bound using Dynamic Latent Variables}
\label{sec:elbo_sde}
Consider two SDEs, starting with the same starting point $\tz_0 = z_0$ at $t=0$,
sharing the same dispersion coefficient $\sigma(z_t, t)$ but
potentially different initial distributions --- $z_0 \sim p_0(z_0)$ for the model,
with path measure $\mathbb{P}_\theta$, and $z_0 \sim q_0(z_0)$ for the variational
posterior, with path measure $\mathbb{Q}$:
\begin{align}
d\tz_t &= h_\theta(\tz_t, t)dt + \sigma(\tz_t, t)d \tilde w_t,  && \quad \text{(model, path measure $\mathbb{P}_\theta$)} \label{eq:true_latent_dynamics}\\
dz_t &= h_\phi(z_t, t)dt + \sigma(z_t, t)dw_t, && \quad \text{(variational posterior, path measure $\mathbb{Q}$)} \label{eq:approx_latent_dynamics}
\end{align}
Where $\tilde w_t$ and $w_t$ are Wiener processes under corresponding path measures.
The first equation can be viewed as the latent dynamics under the model $h_\theta$
we are interested in learning and the second as the latent dynamics under some
variational approximation to the posterior that can be used to produce samples $z_t$.
Further, let $x_{t_i}$ be observations at time $t_i$ that are assumed to only
depend on the corresponding unobserved latent state $z_{t_i}$, then the ELBO can
be written as
\begin{align}
\ln p_\theta(x_{t_1}, \ldots, x_{t_n}) &\ge \mathbb{E}_{\mathbb{Q}}\!\left[\sum_{i=1}^n \ln p_\theta(x_{t_i} | z_{t_i}) -\ln \frac{q_0(z_0)}{p_0(z_0)} - \frac{1}{2}\int_0^T \|u(z_t, t)\|^2\,dt\right] \label{eq:sde_elbo} \\
& = \mathbb{E}_{\mathbb{Q}}\!\left[\sum_{i=1}^n \ln p_\theta(x_{t_i} | z_{t_i})\right]  - \mathrm{KL}(\mathbb{Q} \| \mathbb{P}_\theta) \label{eq:sde_elbo_kl}
\end{align}
Where $u$ satisfies
\begin{align}
  \label{eq:u_def}
\sigma(z, t) u(z, t) = h_\phi(z, t) - h_\theta(z, t)   
\end{align}
We provide the proof of the above general form in \Cref{sec:proof_sde_elbo}.
Similar to the ELBO for the VAEs \citep{kingma2013auto}, the first term in
\cref{eq:sde_elbo_kl} explains observations given the latent
path and the second term penalizes the mismatch between the variational
and model path distributions. In the following, we focus on the case
of $q_0 = p_0$ and draw attention to the general case when needed.

\subsection{Diffusion Bridge}
\label{sec:diffusion_bridge}
Given two arbitrary points $z_0$ and $z_1$, a diffusion bridge between the two is a random process constrained to start and end at the two given end points. A diffusion bridge can be used to specify the stochastic dynamics of a particle that starts at $z_0$ at $t=0$ and is constrained to land at $z_1$ at $t=1$. Consider a stochastic process starting at $z_0$ with the dynamics specified by \cref{eq:approx_latent_dynamics}. Using Doob's h-transform, the SDE for the end point conditioned diffusion bridge, constrained to end at $z_1$ at time $t=1$ can be written as
\begin{align}
\label{eq:doobs}
dz_t &= [h_\phi(z_t, t) + \sigma(z_t, t)\sigma(z_t, t)^T\nabla_{z_t}\ln p(z_1|z_t)]dt + \sigma(z_t, t)dw_t
\end{align}
where $p(z_1 | z_t)$ is the conditional density for $z_1$ under the original dynamics in \cref{eq:approx_latent_dynamics} and depends on $h_\phi$. Note that a Brownian bridge is a special case of a Diffusion bridge where the dynamics are specified by the standard Brownian motion. Diffusion bridges can be used to construct a stochastic mapping between two distributions by considering the end points  $z_0 \sim p_0(z_0)$ and $z_1 \sim p_1(z_1)$ to be sampled from the two distributions of interest.

%%%%%%%%%%%%%%%%%%%%%%%%%%%%%%%%%%%%%%%%%%%%%%%%%%%%%%%%%%%%%%%%%%%%%
\section{Latent Stochastic Interpolants}
\label{sec:lsi}

\paragraph{Stochastic Interpolants (SI) and their limitation:}
SI \citep{albergo2023stochastic} is a powerful framework for generative modeling,
capable of learning a model that can flexibly transform between two probability
distributions.
Let $x_1 \sim p(x_1)$ be an observation from the data distribution $p(x_1)$ that
we want to sample from.
In SI framework, another distribution $p_0(x_0)$ is chosen as a prior with
samples $x_0 \sim p_0(x_0)$.
Typically, $p_0$ is easy to sample from, e.g. a Gaussian distribution.
A stochastic interpolant $x_t$ is then constructed with the requirement that the
marginal distribution $p_t(x_t)$ of $x_t$ equals $p_0$ at $t=0$ and $p_1$ at $t=1$.
For example, the interpolant
$x_t = (1-t)x_0 + tx_1 + \sqrt{t(1-t)}\epsilon, \epsilon \sim N(0, I)$
satisfies this requirement. 
The velocity field and the score function for the generative model are then
estimated as solutions to particular least squares problems.
The learned velocity field and the score function can then be used to transform
a sample from $p_0$ to produce a sample from $p_1$.
SI requires that the samples $x_0$ and $x_1$ are observed, though $x_1$ could be
an output of a \emph{fixed} model, hence still observed.
We use the term observation space SI to emphasize this.

However, we are interested in jointly learning a generative model in a latent
space to leverage efficiency of low dimensional representations while also
aligning the latents with the generative process. Therefore, we want to jointly
optimize an encoder $p_\theta(z_1| x_1)$ that represents high dimensional
observations in the latent space and a decoder $p_\theta(x_1| z_1)$ that maps a
given latent representation to the observation space, along with the generative
model in latent space.
To use SI, we need to interpolate between a fixed prior $p_0(z_0)$ in the latent
space and the true marginal posterior $p_1(z_1) \equiv \int p(z_1 | x_1)dx_1$.
However, we only have access to the posterior model $p_\theta(z_1| x_1)$ that is
optimized concurrently and is an approximation to the true intractable posterior.
Consequently, we can not directly construct an interpolant in the latent space
that satisfies the requirements of SI.
In the following, we address this issue by deriving Latent Stochastic
Interpolants (LSI), though from an entirely different perspective than is
considered by SI.

\paragraph{Generative model with dynamic latent variables:}
Since we want to jointly learn the generative model in a latent space, we propose a latent variable model where the unobserved latent variables are assumed to evolve in continuous time according to the dynamics specified by an SDE of the form in \cref{eq:true_latent_dynamics}. Let $p_\theta(x_1 | z_1)$ be a parameterized stochastic decoder and $h_\theta$ parameterized drift for \cref{eq:true_latent_dynamics}. Then, the generation process using our model is as following -- first a sample $z_0 \sim p_0(z_0)$ is produced from a prior $p_0(z_0)$, then $z_0$ evolves according to the dynamics specified by \cref{eq:true_latent_dynamics} using $h_\theta$ from $t=0$ to $t=1$ to yield a $z_1$, and finally an observation space sample is produced using the decoder $p_\theta(x_1 | z_1)$.
In theory, we can now utilize the ELBO presented in \cref{sec:elbo_sde} to train this model.
Note that, although the ELBO in \cref{eq:sde_elbo} supports arbitrary number of observations $x_{t_i}$ at arbitrary times $t_i$, in this paper we focus on a single observation $x_1$ at $t = 1$.
The ELBO in \cref{eq:sde_elbo} needs a variational approximation to the posterior $p_\theta(z_t | x_1)$ which can be used to sample $z_t$. This approximation is constructed as another dynamical model specified by the SDE in \cref{eq:approx_latent_dynamics}. Unfortunately, for a general variational approximation specified by an arbitrary $h_\phi$, simulating \cref{eq:approx_latent_dynamics} would lead to significant computational burden for large problems during each training iteration and open the door to additional issues resulting from approximations needed for simulation of the SDE. Instead, we explicitly construct the drift $h_\phi$ in \cref{eq:approx_latent_dynamics} such that $z_t$ can be sampled directly without simulation for any time $t$. Our scheme provides a scalable alternative that allows simulation free efficient training, as is common in the observation space diffusion models.

\paragraph{Variational posterior with simulation free samples:}
Next we construct a variational posterior approximation, that enables easy
sampling of $z_t$ without requiring the simulation of the SDE in \cref{eq:approx_latent_dynamics}.
Let $z_1 \sim p_\theta(z_1 | x_1)$ be a stochastic encoding of the observation $x_1$ providing direct access to $z_1$ at $t=1$.
Next, using the Diffusion Bridge specified by \cref{eq:doobs} we construct a stochastic mapping between the prior $p_0(z_0)$ and the aggregated approximate posterior $\int p_\theta(z_1 |x_1)dx_1$ at $t=1$.
The diffusion bridge, coupled with the encoder $p_\theta(z_1 | x_1)$ yields our approximate posterior $p_\theta(z_t | x_1)$. 
However, $p(z_1|z_t)$ is unknown in general. 
If we additionally assume that $h_\phi(z_t, t) \equiv h_t z_t$ and $\sigma(z_t, t) \equiv \sigma_t$, then the original SDE in \cref{eq:approx_latent_dynamics} becomes linear with  additive noise
\begin{align}
\label{eq:gaussian_sde}
dz_t &= h_tz_tdt + \sigma_tdw_t
\end{align}
It is well known that for linear SDEs of the above form, the transition density $p(z_t | z_s), t > s$ is gaussian $N(z_t; a_{st}z_s, b_{st}I)$ (see \cref{sec:gaussian_conditional}) for some functions $a_{st}, b_{st}$ that depend on $h_t, \sigma_t$. Consequently, we can compute $\nabla_{z_t}\ln p(z_1|z_t)$ for a given $z_t$ as
\begin{align}
\label{eq:grad_log_z1_zt}
    \nabla_{z_t}\ln p(z_1|z_t) &= \frac{a_{t1}(z_1 - a_{t1}z_t)}{b_{t1}}
\end{align}
The transformed SDE in terms of the simplified drift and dispersion coefficients can be expressed as
\begin{align}
\label{eq:doobs_simple}
dz_t &= [h_tz_t + \sigma^2_t\nabla_{z_t}\ln p(z_1|z_t)]dt + \sigma_tdw_t
\end{align}
Further, if we condition on the starting point $z_0$, then the conditional density $p(z_t | z_1, z_0)$ can be expressed as following using the Bayes rule
\begin{align}
\label{eq:gauss_zt_g_z0_z1}
    p(z_t | z_1, z_0) &= \frac{p(z_1 | z_t, z_0)p(z_t | z_0)}{p(z_1 | z_0)} = \frac{p(z_1 | z_t)p(z_t | z_0)}{p(z_1 | z_0)}
\end{align}
where $p(z_1 | z_t, z_0) = p(z_1 | z_t)$ because of the Markov independence assumption inherent in \cref{eq:approx_latent_dynamics}. Note that all the factors on the right are gaussian. It can be shown that the conditional density $p(z_t | z_1, z_0)$ is also gaussian if the transition densities are gaussian and takes the following form
\begin{align}
    \label{eq:gauss_zt_g_z0_z1_full}
p(z_t | z_1, z_0) &= \left({\frac{1}{2 \pi}\frac{b_{01}}{b_{0t}b_{t1}}}\right)^{\frac{d}{2}} \exp\left(-\frac{1}{2}\frac{b_{01}}{b_{0t}b_{t1}}\left\lVert z_t - \frac{b_{0t}a_{t1}z_1 + b_{t1}a_{0t}z_0}{b_{01}}\right\rVert^2\right)
\end{align}
Where $a_{(\cdot)}, b_{(\cdot)}$ are constant or time dependent scalars and $d$ is the dimensionality of $z_t$. Their specific forms depends on the choice of $h_t, \sigma_t$. Refer to \cref{sec:gaussian_conditional} for details. $z_t$ can now be directly sampled without simulating the SDE, given a sample $z_0$ and the encoded observation $z_1$. 
Note that the assumptions made for \cref{eq:gaussian_sde}, while restrictive, do not limit the empirical performance.

\paragraph{Latent stochastic interpolants:}
We can now define latent stochastic interpolants using reparameterization trick
in conjuction with \cref{eq:gauss_zt_g_z0_z1_full} to parameterize $z_t$ as
\begin{align}
\label{eq:lsi_interpolant}
    z_t = \eta_t\epsilon + \kappa_t z_1 + \nu_t z_0, \quad \epsilon \sim N(0, I)
\end{align}
For some functions $\eta_t, \kappa_t, \nu_t$ that depend on $a_{(\cdot)}, b_{(\cdot)}$. Note that $\eta_0 = \eta_1 = 0, \kappa_0 = \nu_1 = 0, \kappa_1 = \nu_0 = 1$ since $z_t$ is sampled from a diffusion bridge with the two end points fixed at $z_0, z_1$. \Cref{eq:lsi_interpolant} specifies a general stochastic interpolant, akin to the proposal in \citep{albergo2023stochastic}, but now in the latent space. If we choose the encoder and decoder to be identity functions, then above can be viewed as an alternative way to construct stochastic interpolants in the observation space. Instead of choosing $h_t, \sigma_t$ first, we can instead choose $\kappa_t, \nu_t$ and infer the corresponding $h_t, \sigma_t$. For example, choosing $\kappa_t=t, \nu_t=1-t$ leads to $\sigma_t=\sigma$, a constant, and we arrive at the following
\begin{align}
    z_t &= \sigma \sqrt{t(1-t)}\epsilon + tz_1 + (1-t)z_0, \quad \epsilon \sim N(0, I)
\end{align}
See \cref{sec:formulation_linear_app} for a detailed derivation.
We use the above form for all the experiments in the paper.
Further, if $p_0(z_0)$ is chosen to be a standard gaussian then the interpolant simplifies to $z_t = tz_1 + \sqrt{(1-t)(\sigma^2 t + 1 - t)}z_0$ (\cref{sec:gaussian_z_0}).
With the above interpolants, we can now define the ELBO and optimize it efficiently with simulation free samples $z_t$. We also derive the expressions for variance preserving choices of $\kappa_t = \sqrt{t}, \eta_t^2 + \nu_t^2 = 1-t$ in \cref{sec:formulation_var_preserve_app}, however we do not explore this interpolant empirically.

\paragraph{Constructing training objective using ELBO (\cref{eq:sde_elbo}):}
We first define $u(z_t, t)$ using \cref{eq:doobs_simple} as
\begin{align}
\label{eq:u}
u(z_t, t) &= \sigma_t^{-1}[h_tz_t + \sigma^2_t\nabla_{z_t}\ln p(z_1|z_t) - h_\theta(z_t, t)]
\end{align}
For the general latent stochastic interpolant $z_t = \eta_t\epsilon + \kappa_t z_1 + \nu_t z_0$ (\cref{eq:lsi_interpolant}), we show that $u(z_t, t)$ takes the following form
\begin{align}
\label{eq:general_u}
u(z_t, t) &= \sigma_t^{-1}\left[\left(\frac{d\eta_t}{dt}-\frac{\sigma_t^2}{2\eta_t }\right)\epsilon + \frac{d\kappa_t}{dt}z_1 + \frac{d\nu_t}{dt}z_0 - h_\theta(z_t, t)\right]
\end{align}
See \cref{sec:general_training_obj_app} for the proof. This $u(z_t, t)$ can be substituted into the ELBO in \cref{eq:sde_elbo} to construct a training objective. For example, 
with the choices $\kappa_t=t, \nu_t=1-t$, we get
\begin{align}
u(z_t, t) &= \sigma^{-1}\left[-\sigma\sqrt{\frac{t}{1-t}}\epsilon + z_1 - z_0 - h_\theta(z_t, t)\right]
\end{align}
See \cref{sec:formulation_linear_app} for details. We write a generalized loss based on the ELBO as
\begin{align}
\label{eq:method_gen_elbo}
    \mathbb E_{p(t)p(x_1, z_0)p_\theta(z_1|x_1)p(z_t|z_1, z_0)} \left[-\ln p_\theta(x_1 | z_1) + \frac{\beta_t}{2}\bigg\lVert \sigma\sqrt{\frac{t}{1-t}}\epsilon + z_1 - z_0 - h_\theta(z_t, t)\bigg\rVert^2\right]
\end{align}
Where $\beta_t$ (discussed further in \cref{sec:parameterization}) is a relative
weighting term, similar in spirit to $\beta$-VAE
\citep{higgins2017beta, alemi2018fixing}, allowing empirical re-balancing for
metrics of interest, e.g. FID.
Above loss is reminiscent of the SI training objective, but with an additional
reconstruction term and the interpolants $z_t$ arising from the variational
posterior. We use this training objective for all the experiments,
and optimize it using stochastic gradient descent to jointly
train all three components -- encoder $p_\theta(z_1|x_1)$,
decoder $p_\theta(x_1|z_1)$ and latent SI model $h_\theta(z_t, t)$.
Note that we choose $p_\theta(x_1|z_1)$ to be a conditional gaussian in all
experiments, resulting in a simple $L_2$ decoder loss.

\paragraph{Observation-space stochastic interpolants:}
To elucidate the connection with observation-space SI \citep{albergo2023stochastic}
we derive the corresponding training objective in our framework, yielding:
\begin{align}
    \label{eq:obs_elbo_linear_main}
    \mathbb E_{p(t)p(x_1, x_0)p(x_t|x_1, x_0)} \left[\frac{\beta_t}{2}\bigg\lVert \sigma\sqrt{\frac{t}{1-t}}\epsilon + x_1 - x_0 - h_\theta(x_t, t)\bigg\rVert^2\right]
\end{align}
where $\beta_t$ has the same interpretation as in \cref{eq:method_gen_elbo},
with $\beta_t = \sigma^{-2}$ corresponding to exact ELBO.
See \Cref{sec:obs_space_si} for detailed proof.
Comparing with the LSI loss (\cref{eq:method_gen_elbo}),
the observation-space ELBO is precisely the LSI objective with the
reconstruction term $-\ln p_\theta(x_1 | z_1)$ removed and $z$ replaced by
$x$.
LSI recovers observation-space stochastic interpolants when the encoder and
decoder are identity functions.
All parameterizations (\Cref{sec:parameterization})
and sampling procedures (\Cref{sec:sampling}) apply directly with $z$ replaced
by $x$. Lastly, the likelihood control property of the above objective is
trivially established -- the objective corresponds to
$\mathrm{KL}(\mathbb{Q} \| \mathbb{P}_\theta)$ for $\beta_t = \sigma^{-2}$ and
$\mathrm{KL}(p_1 \| p_\theta) \le \mathrm{KL}(\mathbb{Q} \| \mathbb{P}_\theta)$
(\cref{eq:likelihood_path_kl_bound}), where $p_1$ is the true data distribution
and $p_\theta$ is the data likelihood under the model.

\paragraph{Learnable priors:} When the prior $p_0$ is parameterized
(e.g., $p_\theta(z_0) = \mathcal{N}(\mu_\theta, \Sigma_\theta)$), the default
construction above uses the same learnable prior for both processes
($q_0 = p_\theta$), so $\mathrm{KL}(q_0 \| p_0) = 0$
and the ELBO retains the
same form. The prior parameters are still learned: they affect the distribution
of $z_0$ in the path integral $\mathbb{E}_\mathbb{Q}[\int \|u\|^2\,dt]$, and
gradients flow through $z_0 \sim p_\theta(z_0)$ via the reparameterization
trick. Alternatively, if the variational process uses a fixed reference
$q_0 \neq p_\theta$, the $\mathrm{KL}(q_0 \| p_\theta)$ term appears as an
additional regularizer penalizing deviation from the reference. Same carries
over to the observation-space stochastic interpolants as well.

%%%%%%%%%%%%%%%%%%%%%%%%%%%%%%%%%%%%%%%%%%%%
\section{Parameterization}
\label{sec:parameterization}

Directly using the loss in \cref{eq:method_gen_elbo} leads to high variance in gradients and unreliable training due to the $\sqrt{1-t}$ in the denominator of the second term. Consequently, we consider several alternative parameterizations for the second term, including denoising and noise prediction (see \cref{sec:parameterizations_app} for details). Among the alternatives considered, we found the following parameterization, referred to as \flow{}, to reliably lead to better results and we use it in all our experiments.
\begin{align}
\label{eq:flow_loss}
 \frac{\beta_t}{2}\left\lVert-\sigma\sqrt{t}\epsilon + \sqrt{1-t}(z_1 - z_0) + \sqrt{t}z_t - \hat h_\theta(z_t, t)\right\rVert^2
\end{align}
Where $\hat h_\theta(z_t, t) \equiv \sqrt{t}z_t + \sqrt{1-t}h_\theta(z_t, t)$ and $\beta_t \equiv \beta/(1-t)$ is a time $t$ dependent weighting term, with $\beta$ a constant.
Instead of explicitly using the weights $\beta_t$, due to $1-t$ in the denominator, we consider a change of variable for $t$ with the parametric family $t(s) = 1 - (1-s)^c$ with $s \sim \mathcal U[0, 1]$ uniformly sampled. It can be shown that $p(t) \propto (1 - t)^{\frac{1}{c}-1}$, therefore the change of variable provides the reweighting and we simply set $\beta_t = \beta$, a constant. Empirically, we found that a value of $c=1$ (i.e. a uniform schedule) works the best for all parameterizations during training and sampling, except for \noisepred{} and \denoising{}, which preferred $c \approx 2$ during sampling. $c < 1$ led to degradation in FID. \Cref{fig:t_schedule} in appendix visualizes $t(s)$ for various values of $c$. While the ELBO suggests using $\beta=1/\sigma^2$, we compute the two terms in \cref{eq:method_gen_elbo} as averages and experiment with different weightings. When used with optimizers like Adam or AdamW, $\beta$ can be interpreted as the relative weighting of the gradients from the two terms for the encoder $p_\theta(z_1 | x_1)$. A lower value of $\beta$ leads the encoder to focus purely on the reconstruction and is akin to using a pre-trained encoder-decoder pair as $\beta \rightarrow 0$. A higher value of $\beta$ forces the encoder to adapt its representation for the second term as well. We empirically study the effect of $\beta$ in the experiments.

%%%%%%%%%%%%%%%%%%%%%%%%%%%%%%%%%%%%%%%%%%%%%%%%%%%%%%%%%%%%%%%%%%%%%%%%%%%%%%%%%%%%%%%%
\section{Sampling}
\label{sec:sampling}
For the \flow{} parameterization, the learned drift $\hat h_\theta(z_t, t)$ is related to the original drift $h_\theta(z_t, t)$ as $h_\theta(z_t, t) = (\hat h(z_t, t) - \sqrt{t}z_t)/\sqrt{1 - t}$ (see \cref{sec:detailed_flow_sampler}). We can sample from the model by discretizing the SDE in \cref{eq:true_latent_dynamics}, where $\sigma_t = \sigma$ for the choices of $\kappa_t = t, \nu_t = 1-t$.
However, to derive a flexible family of samplers where we can independently tune the dispersion $\sigma$ without retraining, we exploit Corollary 1 from \cite{singh2024stochastic} to introduce a family of SDEs with the same marginal distributions as that for \cref{eq:true_latent_dynamics}
\begin{align}
    \label{eq:flex_simp_sampler}
    dz_t &= \left[h_\theta(z_t, t) - \frac{(1-\gamma_t^2)\sigma^2}{2}\nabla_{z_t}\ln p_t(z_t)\right]dt + \gamma_t \sigma dw_t
\end{align}
Where $\gamma_t \ge 0$ can be chosen to control the amount of stochasticity introduced into sampling. 
For example, setting $\gamma_t=0$ yields the probability flow ODE for deterministic sampling.
In general, to use \cref{eq:flex_simp_sampler} for $\gamma_t \ne 1$, the score function $\nabla_{z_t} \ln p_t(z_t)$ is needed as well. For the interpolant $z_t = \sigma\sqrt{t(1-t)}\epsilon + tz_1 + (1-t)z_0$, the score can be estimated using
\begin{align}
\label{eq:score_estimator}
    \nabla_{z_t} \ln p_t(z_t) &= -\frac{\mathbb E[\epsilon | z_t]}{\sigma \sqrt{t(1-t)}}
\end{align}
See \cref{sec:general_score_fn} for the proof. However, for Gaussian $z_0$, score can be computed from the drift $h_\theta(z_t, t)$ \citep{singh2024stochastic} as following (see \cref{sec:score_fn} for details)
\begin{align}
\label{eq:cfg_drift}
    \nabla_x \ln p_t(z_t) &= -z_t + t h_\theta(z_t, t)
\end{align}
\Cref{sec:detailed_sampling} provides detailed derivation of samplers for various parameterizations. For classifier free guided sampling \citep{ho2022classifier,xie2024reflected,dao2023flow,zheng2023guided, singh2024stochastic}, we define the guided drift as a linear combination of the conditional drift $h_\theta(z_t, t, c)$ and the  unconditional drift $h_\theta(z_t, t, c=\varnothing)$ as
\begin{align}
    h^{\text{cfg}}(z_t, t, c) \equiv (1+\lambda) h_\theta(z_t, t, c) - \lambda h_\theta(z_t, t, c=\varnothing)
\end{align}
where $\lambda$ is the relative weight of the guidance, $c$ is the conditioning information and $c=\varnothing$ denotes no conditioning.
Note that $\lambda = -1$ corresponds to unconditional sampling, $\lambda = 0$ corresponds to conditional sampling and $\lambda > 0$ further biases towards the modes of the conditional distribution.

%%%%%%%%%%%%%%%%%%%%%%%%%%%%%%%%%%%%%%%%%%%%%%%%%%%%%%%%%%%%%%%%%%%%%
\section{Experiments}
\label{sec:experiments}

\begin{table}[t]
  \caption{%
    \textbf{LSI enables joint learning for SI and cheaper sampling:} The latent space models achieve FID similar to observation space models of comparable size. However, the latent space model \LAT{} has fewer parameters (reported in millions (M)) and FLOPs (reported in Giga (G)), as part of the parameters live in the encoder \ENC{} and the decoder \DEC{}. During sampling, encoder is not used, decoder is used only once, while the latent model \LAT{} is run repeatedly, once for each sampling step. Therefore, FLOP savings from a computationally cheaper latent model accumulate with sampling steps.
  }
  \label{tab:resolution_fid}
  \centering
  \begin{tabular}{lLLLLLL}
    \toprule
    & \multicolumn{2}{c}{\text{FID @ 2K epochs}} & \multicolumn{2}{c}{\text{\# Params (M)}} & \multicolumn{2}{c}{\text{Flops (G)}}\\
    \cmidrule(r){2-3} \cmidrule(r){4-5} \cmidrule(r){6-7}
    Resolution & \text{Latent} & \text{Observ.} & \text{Latent (\EDL{E}{D}{L})} & \text{Observ.} & \text{Latent (\EDL{E}{D}{L})} & \text{Observ.} \\
    \midrule
    $64 \times 64$    & $2.62$ & $2.57$ & $392 (\EDL{5}{5}{382})$ & $398$ & $\EDL{15}{15}{161}$ & $201$ \\
    $128 \times 128$  & $3.12$ & $3.46$ & $392 (\EDL{5}{5}{382})$ & $400$ & $\EDL{59}{59}{327}$ & $466$ \\
    $256 \times 256$  & $3.91$ & $3.87$ & $393 (\EDL{5}{5}{383})$ & $405$ & $\EDL{240}{240}{450}$ & $1288$ \\
    \bottomrule
  \end{tabular}
\end{table}

We evaluate LSI on the standard ImageNet (2012) dataset~\citep{deng2009imagenet,ILSVRC15}.
We train models at various image resolutions and compare their sample quality using the Frechet Inception Distance (FID) metric~\citep{heusel2017gans} for class conditional samples. All models were trained for $1000$ epochs, except for the comparison in \cref{tab:resolution_fid} which reports FID at $2000$ epochs. All results use deterministic sampler, using $\gamma_t=0$, unless otherwise specified.
A key implementation detail to note is that the encoder uses normalization and \texttt{tanh} to bound the scale of the latents. See \cref{sec:add_imagenet_details,sec:arch_details_app} for additional details.

\paragraph{LSI enables joint learning for SI :} While SI doesn't allow latent variables, LSI enables joint learning of Encoder (\ENC{}), Decoder (\DEC{}), and Latent SI models (\LAT{}). In \cref{tab:resolution_fid} we compare FID across various resolutions for LSI models against SI models trained directly in observation (pixel) space. LSI models achieve FIDs similar to the observation space models indicating on par performance in terms of the final FID. Models for both were chosen with similar architecture and number of parameters and trained for 2000 epochs. Reference comparison with other methods is provided in \cref{sec:other_comparison_app}.

\paragraph{LSI enables computationally cheaper sampling:} In \cref{tab:resolution_fid} we also report the parameter counts (in millions) as well as FLOPs (in Giga) for the observation space SI model as well as \ENC{}, \DEC{} and \LAT{} models for the LSI. For the latent \LAT{} model, FLOPs are reported for a single forward pass. First note that the parameters in LSI are partitioned across the encoder \ENC{}, the decoder \DEC{} and the latent \LAT{} models. At sampling time, encoder is not used, decoder is used only once, while the latent model is run multiple times, once for each step of sampling. Therefore, while the overall FLOP count for LSI and Observation space SI models is similar for a single forward pass, sampling with multiple steps becomes significantly cheaper. For example, sampling with 100 steps leads to $29.7\%$ reduction in FLOPs for sampling $128 \times 128$ images and $64.9\%$ for $256 \times 256$ images.

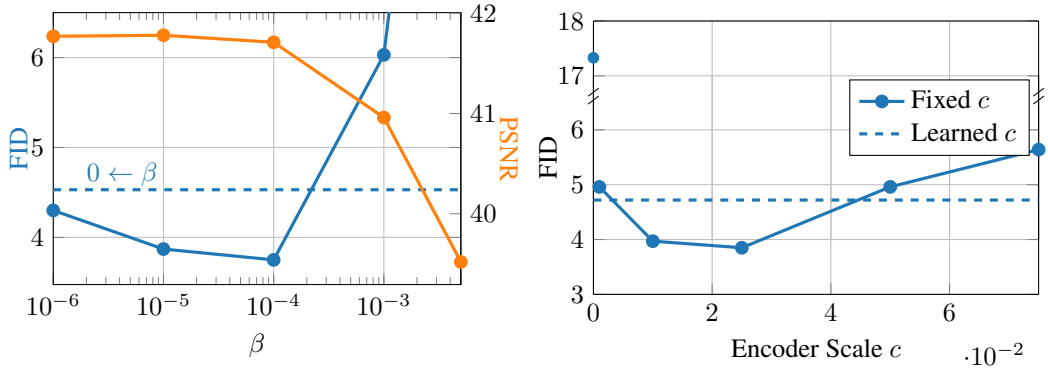
\begin{figure}[t]
\centering
\begin{subfigure}[b]{0.49\linewidth}
\begin{tikzpicture}
    \begin{axis}[
        width=7cm,
        height=5.2cm,
        xlabel={$\beta$},          % Label for the x-axis
        ylabel={FID},          % Label for the y-axis
        ylabel near ticks,
        ylabel style={yshift=-0.2cm, color=sns0},
        % title={Line Plot},   % Title of the plot
        grid=major,          % Add a major grid
        legend pos=north west, % Position of the legend
        xmode=log,
        xmin=1e-06,
        xmax=5e-03,
        ymax=6.5,
    ]
    
    \addplot[
        color=sns0,        % Color of the line
        mark=*,            % Marker style (e.g., *, o, square, triangle)
        mark options={fill=sns0}, % Fill color for the markers
        very thick,
    ] coordinates {
        (1e-06,4.30)
        (1e-05,3.87)
        (0.0001,3.75)
        (0.001,6.03)
        (0.005,12.93)
    };
    % \legend{Data} % Legend entry for the plot

    \addplot [
        color=sns0,      % Color of the line
        dashed,         % Style of the line
        domain=1e-06:5e-03, % Draw across the whole x-axis range
        samples=2,      % Only need 2 samples for a straight line
        very thick,
    ] {4.53}; % Plots y=4

    \node[
        color=sns0,              % Color of the text, matching the line
        above,                  % Position the node above the coordinate
        anchor=east,            % Anchor the text's right edge at the coordinate
        inner sep=2pt,          % Small padding around the text
        %xshift=-2pt            % Optionally shift slightly left from the very edge
    ] 
    at (axis cs:1e-05, 4.7) % Coordinate: (xmax, y=4)
    {$0 \leftarrow \beta$};
    
    \end{axis}

    \begin{axis}[
        width=7cm,
        height=5.2cm,
        axis y line*=right,
        axis x line=none,
        ylabel={PSNR},          % Label for the y-axis
        ylabel near ticks,
        ylabel style={rotate=180, yshift=-0.2cm, color=sns1},
        % title={Line Plot},   % Title of the plot
        % grid=major,          % Add a major grid
        legend pos=north west, % Position of the legend
        xmode=log,
        xmin=1e-06,
        xmax=5e-03,
    ]
    
    \addplot[
        color=sns1,        % Color of the line
        mark=*,            % Marker style (e.g., *, o, square, triangle)
        mark options={fill=sns1}, % Fill color for the markers
        very thick,
    ] coordinates {
        (1e-06,41.77)
        (1e-05,41.78)
        (0.0001,41.71)
        (0.001,40.96)
        (0.005,39.52)
    };
    % \legend{Data} % Legend entry for the plot
    \end{axis}
\end{tikzpicture}
% \caption{\textbf{Effect of $\beta$}}
\label{fig:beta_eval}
\end{subfigure}
\hfill
\begin{subfigure}[b]{0.49\linewidth}
\begin{tikzpicture}
    \begin{groupplot}[
        group style={
            group size=1 by 2,      % 1 column, 2 rows
            vertical sep=-2.2cm,       % No vertical separation to make them touch
            xticklabels at=edge bottom, % Show xticklabels only on the bottom-most plot
        },
        width=7.5cm,                 % Common width for all plots
        xmin=0, xmax=0.075,             % Common x-axis range for both plots
    ]
    
    % TOP PLOT (shows y-values around 50)
    \nextgroupplot[
        height=4.5cm,               % Height for the upper segment
        ymin=14, ymax=18,           % Y-range for upper data
        ytick={17, 18},             % Y-TICKS for this part (for y-grid lines)
        % xtick={0,1,2,3,4,5},        % ADDED: X-TICKS for x-grid lines
        xmajorticks=false,
        % every axis x line/.append style={draw=orange, thick}, % Apply style using the 'every' hook
        axis x line*=top,
        % xticklabels=\empty,         % Ensure no x-tick labels on this top plot
        % axis x line=none,           % Keep x-axis line hidden
        grid=major,                 % Use major grid lines (will show both x and y grids now)
        % title={Plot with Y-Axis Break (Groupplot)}, % Title for the whole figure
    ]
    % Plot for (5,50) - only marks
    \addplot[
        color=sns0,
        mark=*,
        mark options={fill=sns0},
        only marks, % Ensures only the marker is plotted
    ] coordinates {
        (0.0,17.33)
    }; % Semicolon is important

    % BOTTOM PLOT (shows y-values 0-10)
    \nextgroupplot[
        height=4.5cm,               % Height for the lower segment (more visual space)
        ymin=3, ymax=7,            % Y-range for lower data
        ytick={3,4,5,6},       % Ticks for lower part
        % xtick={0,1,2,3,4,5},        % X-ticks visible here (for labels and grid)
        xlabel={Encoder Scale $c$},                 % X-label on the bottom plot
        grid=major,                 % Show major grid lines (both x and y)
        axis x line*=bottom,        % Ensure bottom x-axis line is drawn
        legend cell align={left},
    ]
    % Plot for (0,1)...(4,2)
    \addplot[
        color=sns0,
        mark=*,
        mark options={fill=sns0},
        very thick, % Bolder line
    ] coordinates {
        (0.001,4.96)
        (0.01,3.97)
        (0.025,3.85)
        (0.05,4.96)
        (0.075,5.64)
    }; % Semicolon is important
    \addlegendentry{Fixed $c$} % Legend entry for the plot
    \addplot [
        color=sns0,      % Color of the line
        dashed,         % Style of the line
        domain=0:0.075, % Draw across the whole x-axis range
        samples=2,      % Only need 2 samples for a straight line
        very thick,
    ] {4.72};
    \addlegendentry{Learned $c$} % Legend entry for the plot
    
    \end{groupplot} % End of groupplot environment

    % --- Manually draw break marks on the y-axis ---
    \coordinate (junction_point1) at ($(group c1r2.north west) + (0pt, -8pt)$); % Top-left of bottom plot's axis box
    \coordinate (junction_point2) at ($(group c1r2.north east) + (0pt, -8pt)$); % Top-Right of bottom plot's axis box
    
    % Draw a "Z" or "N" like break symbol at the junction_point
    \draw[black, line width=0.4pt] ($(junction_point1) + (-2pt, -3pt)$) -- ($(junction_point1) + (2pt, 0pt)$); % Semicolon
    \draw[black, line width=0.4pt] ($(junction_point1) + (-2pt, 0pt)$) -- ($(junction_point1) + (2pt, 3pt)$); % Semicolon

    % Draw a "Z" or "N" like break symbol at the junction_point
    \draw[black, line width=0.4pt] ($(junction_point2) + (-2pt, -3pt)$) -- ($(junction_point2) + (2pt, 0pt)$); % Semicolon
    \draw[black, line width=0.4pt] ($(junction_point2) + (-2pt, 0pt)$) -- ($(junction_point2) + (2pt, 3pt)$); % Semicolon

    % --- Add a single, shared y-label for the entire figure ---
    \node[rotate=90, anchor=center] at ($(group c1r1.west)!0.5!(group c1r2.west) + (-0.6cm,0)$) {FID}; % Semicolon

\end{tikzpicture} % Ends TikZ picture
% \caption{\textbf{Effect of encoder noise}}
\label{fig:outscale_eval}
\end{subfigure}
\caption{\textbf{Effect of loss trade-off $\pmb{\beta}$ and encoder noise scale $\pmb{c}$:} In the left panel, we evaluate the effect of loss trade-off weight $\beta$ for $128 \times 128$ models and observe that FID improves with $\beta$, until the degradation in reconstruction quality (PSNR) starts degrading FID. In the right panel, we evaluate the effect of encoder noise scale on FID. We also plot the FID for a model with learned scale as dashed line. A deterministic encoder performs the worst ($c=0$), with FID improving with $c$ until it degrades again. Encoder with learned $c$ (dashed line) is outperformed by fixed $c$ in our experiments.}
\label{fig:beta_outscale_eval}
\end{figure}

\begin{table}[t]
  \caption{%
    \textbf{Joint training helps mitigate capacity shift:} We evaluate the effect of moving first $k$ and last $k$ convolutional blocks from the latent model \LAT{} to encoder and decoder respectively, for $128 \times 128$ resolution models. This results in the overall parameter count staying roughly the same, but the number of FLOPs required for sampling changing significantly. We observe that the model trained with $\beta > 0$ perform better and maintains FID well, in comparison to the independently trained model ($\beta \rightarrow 0$), even when capacity is shifted away from the latent model \LAT{}, resulting in $8.5\%$ reduction in FLOPs for sampling from $k=0$ to $k=6$.
  }
  \label{tab:capacity_shift_fid}
  \centering
  \begin{tabular}{lLLLL}
    \toprule
    $k$ & \text{FID ($\beta > 0$)} & \text{FID ($\beta \rightarrow 0$)} & \# \text{Params. (\EDL{E}{D}{L})} & \text{FLOPs (\EDL{E}{D}{L})}\\
    \midrule
    $0$ & $3.76$ & $4.31$ & $392 (\EDL{5}{5}{382})$ & $\EDL{59}{59}{327}$ \\
    $3$ & $3.91$ & $4.55$ & $389 (\EDL{9}{8}{372})$ & $\EDL{68}{66}{313}$ \\
    $6$ & $3.96$ & $4.87$ & $387 (\EDL{13}{12}{362})$ & $\EDL{75}{73}{299}$ \\
    $9$ & $4.61$ & $4.98$ & $383 (\EDL{16}{16}{351})$ & $\EDL{82}{80}{284}$ \\
    \bottomrule
  \end{tabular}
\end{table}

\paragraph{Joint learning is beneficial:} In \cref{fig:beta_outscale_eval}(left panel) we plot the FID as the weighting term $\beta$ is varied (\cref{eq:flow_loss}). A higher $\beta$ forces the encoder to adapt the latents more for the second term of the loss. We observe that FID improves as $\beta$ increases, going from $4.53$ (for $\beta \rightarrow 0$) to $3.75$ ($\approx 17 \%$ improvement) for $\beta = 0.0001$, indicating that this adaptation is beneficial for the overall performance. Eventually, FID worsens as $\beta$ is increased further. We also plot the reconstruction PSNR for each of these models in orange and observe that increasing $\beta$ essentially trades-off reconstruction quality with generative performance. For too large a $\beta$, poor reconstruction quality leads to worsening FID. The dashed line indicates the performance when the encoder-decoder are trained independently of the latent model, limit of $\beta \rightarrow 0$. We implement it as a stop gradient operation in implementation, where the gradients from the second term of the loss are not backpropagated into $z_1$. To further assess the benefits of joint training, in \cref{tab:capacity_shift_fid} we compare the FIDs between jointly trained model ($\beta >0$) and independently trained model ($\beta \rightarrow 0$) as parameters are shifted from the latent model \LAT{} to the encoder \ENC{} and decoder \DEC{} models, by moving first $k$ and last $k$ convolutional blocks from the latent model to the encoder and the decoder respectively. While this keeps the total parameter count roughly the same, the number of FLOPs required for sampling changes significantly. The jointly trained model performs better and maintains FID well even when capacity shifts away from the latent model, resulting in $8.5\%$ reduction in FLOPs required for sampling from $k=0$ to $k=6$.

\paragraph{Encoder noise scale affects performance:} The stochasticity of the encoder $p_\theta(z_1 | x)$ has a significant impact on the performance. We parameterize the encoder as a conditional Gaussian $N(z_1; \mu_\theta(x), \Sigma_\theta(x))$ where $\Sigma(x)$ is assumed to be diagonal. We experimented with a purely deterministic encoder ($\Sigma_\theta(x) = 0$), learned $\Sigma_\theta(x)$ and constant noise $\Sigma_\theta(x)=cI$. In \cref{fig:beta_outscale_eval}(right panel) we plot FID as the encoder output stochasticity $c$ is varied. Dashed line indicates performance with learned $\Sigma_\theta(x)$. A deterministic encoder ($c=0$) performs poorly. FID improves as the noise scale $c$ is increased, until eventually it degrades again. While learned  $\Sigma_\theta(x)$ (dashed line) performs well, fixed $c$ models achieved higher FID.

\paragraph{\pmb{\flow{}} parameterization performs better than alternatives:} In \cref{tab:parameterization_fid} we compare different parameterizations discussed in \cref{sec:parameterization} and \cref{sec:parameterizations_app}. The \flow{} parameterization consistently led to better FID. Both \origflow{} and \noisepred{} parameterizations exhibited higher variance gradients and noisy optimization. While \denoising{} parameterization resulted in less noisy training, \flow{} parameterization led to fastest improvement in FID.

\paragraph{LSI supports diverse $p_0$:} In \cref{tab:prior_fid} we report FID achieved by LSI using different prior $p_0(z_0)$ distributions. While Gaussian $p_0$ performs the best, other choices for $p_0$ yield competitive results indicating that LSI retains one of the key strengths of SI -- support for diverse $p_0$ distributions. See \cref{sec:choice_of_priors} for additional details. To allow flexible sampling using \cref{eq:flex_simp_sampler}, we modified latent SI model to output extra output channels and augmented the loss with another term to estimate $\mathbb E[\epsilon | z_t]$. \Cref{eq:score_estimator} was used to compute the score and sample with the deterministic sampler using $\gamma_t = 0$.

\begin{figure}[t]
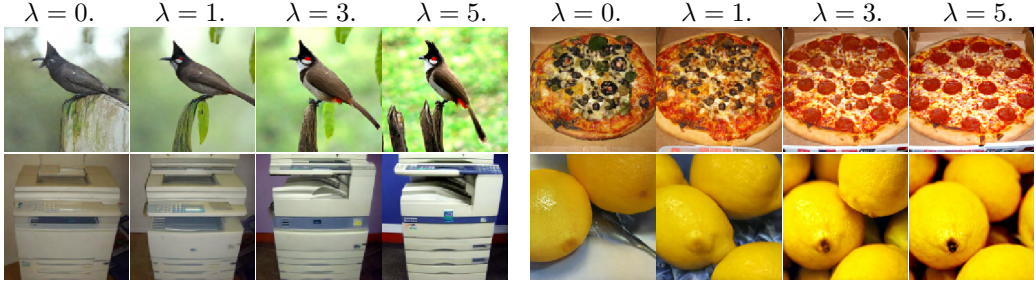

    \centering
    \begin{tikzpicture}
        % Define the width of each image (adjust as needed)
        \def\imagewidth{0.48\textwidth}

        % --- Coordinates for Spacing ---
        \def\newxdist{7.0} % Horizontal distance for the second column
        \def\firstrowy{1.5} % Y-coordinate for the center of the first row images
        \def\rowspacing{1.7} % Vertical distance between the centers of rows (adjust if needed)

        % --- Place images in a 5x2 grid ---

        % Row 1
        \node (img11) at (0, \firstrowy) {\includegraphics[width=\imagewidth]{\imagepath cfg_16_115.pdf}};
        \node (img12) at (\newxdist, \firstrowy) {\includegraphics[width=\imagewidth]{\imagepath cfg_963_18.pdf}};

        % Row 2
        \node (img21) at (0, \firstrowy - \rowspacing) {\includegraphics[width=\imagewidth]{\imagepath cfg_713_9.pdf}};
        \node (img22) at (\newxdist, \firstrowy - \rowspacing) {\includegraphics[width=\imagewidth]{\imagepath cfg_951_37.pdf}};

        % --- Add Labels Above Sub-Images in the First Row ---
        % (This part remains the same as before)
        \coordinate (img11_sub1_top) at ($(img11.north west)!0.125!(img11.north east)$);
        \coordinate (img11_sub2_top) at ($(img11.north west)!0.375!(img11.north east)$);
        \coordinate (img11_sub3_top) at ($(img11.north west)!0.625!(img11.north east)$);
        \coordinate (img11_sub4_top) at ($(img11.north west)!0.875!(img11.north east)$);

        \node[above=-5pt of img11_sub1_top] {$\lambda=0.$}; % Replace 1.1
        \node[above=-5pt of img11_sub2_top] {$\lambda=1.$}; % Replace 1.2
        \node[above=-5pt of img11_sub3_top] {$\lambda=3.$}; % Replace 1.3
        \node[above=-5pt of img11_sub4_top] {$\lambda=5.$}; % Replace 1.4

        \coordinate (img12_sub1_top) at ($(img12.north west)!0.125!(img12.north east)$);
        \coordinate (img12_sub2_top) at ($(img12.north west)!0.375!(img12.north east)$);
        \coordinate (img12_sub3_top) at ($(img12.north west)!0.625!(img12.north east)$);
        \coordinate (img12_sub4_top) at ($(img12.north west)!0.875!(img12.north east)$);

        \node[above=-5pt of img12_sub1_top] {$\lambda=0.$}; % Replace 2.1
        \node[above=-5pt of img12_sub2_top] {$\lambda=1.$}; % Replace 2.2
        \node[above=-5pt of img12_sub3_top] {$\lambda=3.$}; % Replace 2.3
        \node[above=-5pt of img12_sub4_top] {$\lambda=5.$}; % Replace 2.4

    \end{tikzpicture}
    \caption{\textbf{LSI supports CFG sampling.} Class conditional samples are visualized with increasing guidance weight $\lambda$ leading to more typical samples for the class. See text for details.}
    \label{fig:cfg_sampling}
\end{figure}

\begin{figure}[t]
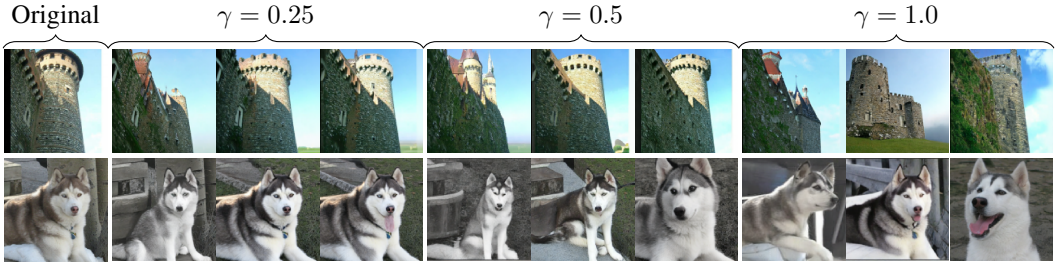
 % htbp are placement specifiers (here, top, bottom, page)
    \centering % Center the figure content

    \begin{tikzpicture}[
        imgnode/.style={inner sep=0pt, outer sep=0pt}, % Style for image nodes to remove padding
        brace_style/.style={decorate, decoration={brace, amplitude=5pt}} % Style for upward braces, reduced amplitude
      ]
        % Node for the first PDF (first row) - Index 6
        \node[imgnode] (img1) {\includegraphics[width=\textwidth, page=1]{\imagepath gamma_grid_6.pdf}};

        % --- Labels ---
        % Labels are positioned with their baseline (anchor=south) 4pt above img1.north.
        % This accounts for a 3pt brace amplitude + 1pt gap.
        \newcommand{\labelYShift}{4pt} % 3pt brace amplitude + 1pt gap

        \node[anchor=south, yshift=\labelYShift] at ($(img1.north west)!0.05!(img1.north east)$) {Original};      % Centered over the first 10% (1 image)
        \node[anchor=south, yshift=\labelYShift] at ($(img1.north west)!0.25!(img1.north east)$) {$\gamma=0.25$}; % Centered over 10%-40% (3 images)
        \node[anchor=south, yshift=\labelYShift] at ($(img1.north west)!0.55!(img1.north east)$) {$\gamma=0.5$};  % Centered over 40%-70% (3 images)
        \node[anchor=south, yshift=\labelYShift] at ($(img1.north west)!0.85!(img1.north east)$) {$\gamma=1.0$};  % Centered over 70%-100% (3 images)

        % --- Braces ---
        % Braces are drawn directly along the top edge of the image (img1.north), pointing upwards.
        % The points for the brace path are on img1.north.
        \draw [brace_style]
            ($(img1.north west)$) -- ($(img1.north west)!0.1!(img1.north east)$);

        \draw [brace_style]
            ($(img1.north west)!0.1!(img1.north east)$) -- ($(img1.north west)!0.4!(img1.north east)$);

        \draw [brace_style]
            ($(img1.north west)!0.4!(img1.north east)$) -- ($(img1.north west)!0.7!(img1.north east)$);

        \draw [brace_style]
            ($(img1.north west)!0.7!(img1.north east)$) -- ($(img1.north east)$);

        % --- Subsequent Images ---
        % Nodes for the subsequent PDFs, positioned with uniform spacing.
        % Using anchor=north and below=<distance> of <previous_node>.south for consistent visual gaps.
        \newcommand{\rowspacing}{0.1em} % Define uniform row spacing - REDUCED for closer rows

        % Index 7
        \node[imgnode, anchor=north, below=\rowspacing of img1.south] (img2) {\includegraphics[width=\textwidth, page=1]{\imagepath gamma_grid_126.pdf}};

    \end{tikzpicture}

    \caption{\textbf{LSI supports flexible sampling.} We demonstrate inversion of an `Original' image, using reverse probability flow ODE (similar to DDIM inversion), followed by forward stochastic sampling to yield samples similar to it, with diversity increasing with $\gamma$ (\cref{eq:flex_simp_sampler}). See text for details.}
    \label{fig:gamma_grid}
\end{figure}

\paragraph{LSI supports flexible sampling:} In \cref{fig:cfg_sampling} and \cref{fig:gamma_grid} we qualitatively demonstrate flexible sampling with LSI model for popular use cases. \Cref{fig:cfg_sampling} demonstrates compatibility of classifier free guidance (CFG) with LSI, using \cref{eq:cfg_drift}. Increasing guidance weight $\lambda$ results in more typical samples. First $z_0$ is sampled from $p_0(z_0)$, Gaussian in this example, following which \cref{eq:flex_simp_sampler} is simulated forward in time, using class conditional drift with different guidance weights $\lambda$.  In \cref{fig:gamma_grid} a given `Original' image (shown leftmost) is first encoded to yield it's representation $z_1$, which is then inverted by simulating probability flow ODE (setting $\gamma_t = 0$ in \cref{eq:flex_simp_sampler}) backward in time from $t=1$ to $t=0$, yielding $z_0$ (similar to DDIM inversion \citep{song2020ddim}). Using this $z_0$ as starting point, \cref{eq:flex_simp_sampler} is simulated forward is time using $\gamma_t \equiv \gamma(1-t)$ for  different values of $\gamma$. We show three samples for each value of $\gamma$ and observe increasing diversity with increasing $\gamma$.  See \cref{sec:sampling_details_app} for additional details and results.

\begin{table}[t]
\parbox{.48\linewidth}{
    \centering
    \caption{
        \textbf{Effect of parameterization:} We compare various parameterization schemes at $128 \times 128$ resolution. \flow{} parameterization performs better against the alternatives.}
    \label{tab:parameterization_fid}
    \begin{tabular}{lL}
    \toprule
    Parameterization & \text{FID @1K epochs}\\
    \midrule
    \origflow{} & $4.56$ \\
    \noisepred{} & $4.73$ \\
    \denoising{} & $4.28$ \\
    \flow{}  & $3.76$ \\
    \bottomrule
    \end{tabular}
}
\hfill
\parbox{.48\linewidth}{
    \centering
    \caption{\textbf{LSI supports diverse $\pmb{p_0}$:} LSI retains one of the key strengths of SI -- support for arbitrary $p_0$ distribution. Different $p_0$ achieve competetive FID for $128 \times 128$ resolution model.}
    \label{tab:prior_fid}
    \begin{tabular}{lL}
    \toprule
    $p_0$ & \text{FID @1K epochs}\\
    \midrule
    Uniform & $4.81$ \\
    Laplacian & $4.45$ \\
    Gaussian & $3.76$ \\
    \midrule
    Gaussian Mixture & 4.26 \\
    \bottomrule
    \end{tabular}
}
\end{table}

%%%%%%%%%%%%%%%%%%%%%%%%%%%%%%%%%%%%%%%%%%%%%%%%%%%%%%%%%%%%%%%%%%%%%
\section{Related Work}

Latent Stochastic Interpolants (LSI) draw from insights in diffusion models,
latent variable models, and continuous-time generative processes.
We discuss key works from these areas in the following.

\noindent
\textbf{Diffusion Models:} Diffusion models, originating from foundational work
on score matching \citep{vincent2011connection,song2019generative} and early
variational formulation \citep{sohl2015deep}, gained prominence with Denoising
Diffusion Probabilistic Models (DDPMs) \citep{ho2020denoising}.
Subsequent improvements focused on architectural choices and learned variances
\citep{nichol2021improved}, faster sampling via Denoising Diffusion Implicit
Models (DDIMs) \citep{song2020ddim}, progressive distillation
\citep{salimans2022progressive}, and powerful conditional generation through
techniques like classifier-free guidance \citep{ho2022classifier}.
Further exploration of the design space
\citep{karras2022elucidating,karras2024analyzing} has lead to highly performant
models.
More recently, diffusion inspired consistency models \citep{song2023consistency}
have emerged, offering efficient generation. LSI complements these with a
flexible method for jointly learning in a latent space using richer prior distributions.

\noindent
\textbf{Latent Variable Models and Expressive Priors:} Variational Autoencoders (VAEs)
\citep{kingma2013auto,rezende2014stochastic} learn a compressed representation $z$
of data $x$, but are limited by the expressiveness of the prior $p(z)$
(NVAE \citep{vahdat2020NVAE}, LSGM \citep{vahdat2021score}), as they typically
use simple priors (e.g., isotropic Gaussian).
LSI addresses this by jointly learning a flexible generative process in the
latent space, enabling powerful transformations of the simple prior.
Early work \citep{sohl2015deep} derived ELBO for discrete time diffusion models,
while Variational Diffusion Models (VDM) \citep{kingma2021variational} interpret
diffusion models as a specific type of VAE with Gaussian noising process.
In contrast, while LSI also optimizes an ELBO, it allows for a broader choice of
the prior $p(z_0)$ and the transforms mapping the prior to the learned aggregated
posterior. Our work is similar in spirit to models like NVAE, which employed
deep hierarchical latent representations, and LSGM, which proposed training
score-based models in the latent space of a VAE, but offers a flexible framework
similar to SI allowing a rich family of priors and latent space dynamics. 
Note that LDM \citep{rombach2022high} train a diffusion generative model in the
latent space of a \emph{fixed} encoder-decoder pair -- making their latents
actually \emph{observed} from the point of view of generative modeling.

\noindent
\textbf{Continuous-Time Generative Processes:} While diffusion models have been
formulated and studied using continuous time dynamics
\citep{song2020score,song2020ddim,kingma2021variational,vahdat2021score}, their
relation to Continuous Normalizing Flows (CNFs)
\citep{chen2018neural,grathwohl2018scalable} offers another perspective on
continuous-time transformations. Early training challenges with the CNFs have
been addressed by newer methods like Flow Matching (FM)
\citep{lipman2022flow,xu2022poisson}, Conditional Flow Matching (CFM)
\citep{neklyudov2023action,tong2023improving}, and Rectified Flow
\citep{liu2022flow}.
These approaches propose simulation-free training by regressing vector fields of
fixed conditional probability paths.
However, likelihood control is typically not possible
\citep{albergo2023stochastic}, consequently extension to jointly learning in
latent space is ill-specified. In contrast, LSI optimizes an ELBO, offering
likelihood control along with joint learning in a latent space.
Stochastic Interpolants (SI) \citep{albergo2023stochastic} provides a unifying
perspective on generative modeling, capable of bridging \emph{any} two
probability distributions via a continuous-time stochastic process, encompassing
aspects of both flow-based and diffusion-based methods.
While SI formulates learning the velocity field and score function directly in
the observation space using pre-specified stochastic interpolants, LSI arrives
at a similar objective in the latent space, as part of the ELBO, from the
specific choices of the approximate variational posterior.
LSI reduces to SI when encoder and decoder are chosen to be Identity functions.
SI is related to the Optimal Transport and the Schrödinger Bridge problem (SBP)
which have been explored as a basis for generative modeling
\citep{debortoli2021generative,wang2021diffusion_sbp,shi2023diffusion}.
While LSI learns a transport, its primary objective is data log-likelihood
maximization via the ELBO, rather than solving a specific OT or SBP.

%%%%%%%%%%%%%%%%%%%%%%%%%%%%%%%%%%%%%%%%%%%%%%%%%%%%%%%%%%%%%%%%%%%%%
\section{Conclusion}
\label{sec:conclusion}

In this paper, we introduced Latent Stochastic Interpolants (LSI), generalizing Stochastic Interpolants to enable joint end-to-end training of an encoder, a decoder, and a generative model operating entirely within the learned latent space. LSI overcomes the limitation of simple priors of the normal diffusion models and mitigates the computational demands of applying SI directly in high-dimensional observation spaces, while preserving the generative flexibility of the SI framework.
LSI leverage SDE-based Evidence Lower Bound to offer a principled approach for optimizing the entire model. We validate the proposed approach with comprehensive experimental studies on standard ImageNet benchmark. Our method offers scalability along with a unifying perspective on continuous-time generative models with dynamic latent variables. However, to achieve scalable training, our approach makes simplifying assumptions for the variational posterior approximation. While restrictive, and common with other methods, these assumptions do not seem to limit the empirical performance.

%%%%%%%%%%%%%%%%%%%%%%%%%%%%%%%%%%%%%%%%%%%%%%%%%%%%%%%%%%%%%%%%%%%%%
\section*{Acknowledgments}

We would like to thank Kevin J. Shih and Ian Fischer for proofreading early
drafts of this manuscript and providing valuable feedback.

%%%%%%%%%%%%%%%%%%%%%%%%%%%%%%%%%%%%%%%%%%%%%%%%%%%%%%%%%%%%%%%%%%%%%
\section*{Reproducibility statement}

We have included detailed proofs of all the key theoretical results in the appendix.
\Cref{sec:experiments,sec:add_imagenet_details} provide key training and evaluation
setup details. \Cref{sec:arch_details_app} provides the necessary architecture details
to reproduce the models used in the experiments. \Cref{sec:sampling_details_app}
provides additional sampling setup details.



\bibliography{references.bib}
\bibliographystyle{iclr2026_conference}

\clearpage

\appendix

\section*{Appendix}

\section{Proof of ELBO for dynamic latent variables (\texorpdfstring{\cref{eq:sde_elbo}}{SDE ELBO})}
\label{sec:proof_sde_elbo}

We provide a self-contained proof of the variational lower bound stated in
\cref{eq:sde_elbo}. The proof is based on the approach of
\citet{li2020scalable}, but in a more general form.
We first establish the path-space KL divergence between two diffusion processes
via Girsanov's theorem (Theorem~\ref{thm:path_kl}), then use it to derive the
ELBO (Theorem~\ref{thm:elbo}). Let $\mathbb{P}_\theta$ be the path measure under
the model as in \cref{eq:true_latent_dynamics} and let $\mathbb{Q}$ be the path
measure under the variational posterior process approximation as in
\cref{eq:approx_latent_dynamics}.

\begin{theorem}[Path-space KL via Girsanov's theorem]
\label{thm:path_kl}
Consider two SDEs, starting at the same starting point $\tz_0 = z_0$, sharing the same dispersion coefficient $\sigma(z_t, t)$ but
potentially different initial distributions --- $z_0 \sim q_0(z_0)$ for
$\mathbb{Q}$ and $z_0 \sim p_0(z_0)$ for $\mathbb{P}_\theta$:
\begin{align}
    d\tz_t &= h_\theta(\tz_t, t)\,dt + \sigma(\tz_t, t)\,dw_t, \qquad \text{(model, path measure $\mathbb{P}_\theta$)} \tag{eq.~\ref{eq:true_latent_dynamics}} \\
    dz_t &= h_\phi(z_t, t)\,dt + \sigma(z_t, t)\,dw_t, \qquad \text{(variational, path measure $\mathbb{Q}$)} \tag{eq.~\ref{eq:approx_latent_dynamics}}
\end{align}
Define $u(z_t, t)$ by $\sigma(z_t, t)\,u(z_t, t) = h_\phi(z_t, t) - h_\theta(z_t, t)$ (\cref{eq:u_def}). Then:
\begin{align}
\label{eq:path_kl}
    \mathrm{KL}(\mathbb{Q} \| \mathbb{P}_\theta) = \mathrm{KL}(q_0 \| p_0) + \frac{1}{2}\,\mathbb{E}_{\mathbb{Q}}\!\left[\int_0^1 \|u(z_t, t)\|^2\,dt\right]
\end{align}
When $q_0 = p_0$ (both processes share the same initial distribution,
potentially learnable), the initial KL vanishes and the path-space KL reduces to
the dynamics mismatch alone.
\end{theorem}

\begin{proof}
The full path measure factorizes as the initial distribution times the
conditional path measure given the initial state. The Radon-Nikodym derivative
therefore decomposes as
\begin{align}
    \frac{d\mathbb{Q}}{d\mathbb{P}_\theta}(\mathcal{Z}) = \frac{q_0(z_0)}{p_0(z_0)} \cdot \frac{d\mathbb{Q}^{z_0}}{d\mathbb{P}_\theta^{z_0}}(\mathcal{Z})
\end{align}
where $\mathbb{Q}^{z_0}$ and $\mathbb{P}_\theta^{z_0}$ denote the conditional
path measures given the initial state $z_0$. Next we use Girsanov's theorem to
evaluate the second factor.
Under $\mathbb{Q}$, the process satisfies 
$dz_t = h_\phi\,dt + \sigma\,dw_t^{\mathbb{Q}}$ where $w_t^{\mathbb{Q}}$ is a
standard Brownian motion.
Define the $\mathbb{P}_\theta$-Brownian motion via $dw_t^{\mathbb{P}_\theta} = \sigma^{-1}(dz_t - h_\theta\,dt)$. Substituting the $\mathbb{Q}$-dynamics for $dz_t$:
\begin{align}
    dw_t^{\mathbb{P}_\theta} = dw_t^{\mathbb{Q}} + u(z_t, t)\,dt
\end{align}
That is, under $\mathbb{Q}$, the process $w_t^{\mathbb{P}_\theta}$ acquires a drift $u_t$. By Girsanov's theorem:
\begin{align}
    \frac{d\mathbb{Q}^{z_0}}{d\mathbb{P}_\theta^{z_0}} = \exp\!\left(\int_0^1 u_t^T\,dw_t^{\mathbb{P}_\theta} - \frac{1}{2}\int_0^1 \|u_t\|^2\,dt\right)
\end{align}
Substituting $dw_t^{\mathbb{P}_\theta} = dw_t^{\mathbb{Q}} + u_t\,dt$ and
combining with the initial density ratio:
\begin{align}
    \ln \frac{d\mathbb{Q}}{d\mathbb{P}_\theta} = \ln \frac{q_0(z_0)}{p_0(z_0)} + \int_0^1 u_t^T\,dw_t^{\mathbb{Q}} + \frac{1}{2}\int_0^1 \|u_t\|^2\,dt
\end{align}
Taking the expectation under $\mathbb{Q}$:
\begin{align}
    \mathrm{KL}(\mathbb{Q} \| \mathbb{P}_\theta) = \underbrace{\mathbb{E}_{q_0}\!\left[\ln \frac{q_0(z_0)}{p_0(z_0)}\right]}_{\mathrm{KL}(q_0 \| p_0)} + \underbrace{\mathbb{E}_{\mathbb{Q}}\!\left[\int_0^1 u_t^T\,dw_t^{\mathbb{Q}}\right]}_{= \, 0} + \frac{1}{2}\,\mathbb{E}_{\mathbb{Q}}\!\left[\int_0^1 \|u_t\|^2\,dt\right]
\end{align}
The It\^o integral $\int_0^1 u_t^T\,dw_t^{\mathbb{Q}}$ is a martingale under
$\mathbb{Q}$ (under the standard integrability condition
$\mathbb{E}_{\mathbb{Q}}[\int_0^1 \|u_t\|^2\,dt] < \infty$),
and the expectation of a martingale starting at zero is zero.
\end{proof}

\begin{remark}
\Cref{thm:path_kl} applies to any two diffusion processes
sharing the same dispersion, regardless of whether the state space represents
latent variables or observations. In particular, it applies directly in
observation space (with $z$ replaced by $x$).
\end{remark}

\begin{remark}[Learnable prior]
When the prior $p_0$ is parameterized
(e.g., $p_\theta(z_0) = \mathcal{N}(\mu_\theta, \Sigma_\theta)$), the natural
construction uses the same learnable prior for both processes
($q_0 = p_\theta$), so $\mathrm{KL}(q_0 \| p_0) = 0$ and the ELBO retains the
same form. The prior parameters are still learned: they affect the distribution
of $z_0$ in the path integral $\mathbb{E}_\mathbb{Q}[\int \|u\|^2\,dt]$, and
gradients flow through $z_0 \sim p_\theta(z_0)$ via the reparameterization
trick. Alternatively, if the variational process uses a fixed reference
$q_0 \neq p_\theta$, the $\mathrm{KL}(q_0 \| p_\theta)$ term appears as an
additional regularizer penalizing deviation from the reference.
\end{remark}

\begin{theorem}[ELBO for dynamic latent variables, \cref{eq:sde_elbo,eq:u_def}]
\label{thm:elbo}
Under the setup of \Cref{thm:path_kl}, with potentially learnable initial
distributions $p_0$ and $q_0$, let $x_{t_i}$ be observations at times
$t_i \in [0, 1]$, $i = 1, \ldots, n$, assumed to depend on the latent state
only through $z_{t_i}$, i.e., $p_\theta(x_{t_i} | \mathcal{Z}) = p_\theta(x_{t_i} | z_{t_i})$.
Then:
\begin{align}
\label{eq:elbo_restated}
    \ln p_\theta(x_{t_1}, \ldots, x_{t_n}) & \ge \mathbb{E}_{\mathbb{Q}}\!\left[\sum_{i=1}^n \ln p_\theta(x_{t_i} | z_{t_i}) -\ln \frac{q_0(z_0)}{p_0(z_0)} - \frac{1}{2}\int_0^1 \|u(z_t, t)\|^2\,dt\right] \\
    & = \mathbb{E}_{\mathbb{Q}}\!\left[\sum_{i=1}^n \ln p_\theta(x_{t_i} | z_{t_i})\right]  - \mathrm{KL}(\mathbb{Q} \| \mathbb{P}_\theta)
\end{align}
When $q_0 = p_0$, the second term on the right vanishes and we recover the
special case of above as stated in \cite{li2020scalable}.
\end{theorem}

\begin{proof}
Under the model, the latent path evolves according to
$\mathbb{P}_\theta$ (\cref{eq:true_latent_dynamics}) and observations are
generated conditionally at each time $t_i$. The marginal likelihood is obtained
by integrating over all latent paths:
\begin{align}
    p_\theta(x_{t_1}, \ldots, x_{t_n}) = \int \prod_{i=1}^n p_\theta(x_{t_i} | z_{t_i})\; d\mathbb{P}_\theta(\mathcal{Z})
\end{align}
Since both $\mathbb{P}_\theta$ and $\mathbb{Q}$ share the same dispersion and
initial distribution, they are mutually absolutely continuous (by Girsanov's
theorem, under standard regularity). We can therefore re-express the integral
using the variational path measure $\mathbb{Q}$ (\cref{eq:approx_latent_dynamics}) as a proposal:
\begin{align}
    p_\theta(x_{t_1}, \ldots, x_{t_n}) = \mathbb{E}_{\mathbb{Q}}\!\left[\prod_{i=1}^n p_\theta(x_{t_i} | z_{t_i}) \cdot \frac{d\mathbb{P}_\theta}{d\mathbb{Q}}(\mathcal{Z})\right]
\end{align}
Taking the logarithm of both sides and using Jensen's inequality:
\begin{align}
    \ln p_\theta(x_{t_1}, \ldots, x_{t_n}) \ge \mathbb{E}_{\mathbb{Q}}\!\left[\sum_{i=1}^n \ln p_\theta(x_{t_i} | z_{t_i}) + \ln \frac{d\mathbb{P}_\theta}{d\mathbb{Q}}(\mathcal{Z})\right]
\end{align}
From the proof of \Cref{thm:path_kl} (with $q_0 = p_0$, so the initial
density ratio cancels), the log Radon-Nikodym derivative expressed in terms of
the $\mathbb{Q}$-Brownian motion is:
\begin{align}
    \ln \frac{d\mathbb{P}_\theta}{d\mathbb{Q}}(\mathcal{Z}) = -\ln \frac{q_0(z_0)}{p_0(z_0)} -\int_0^1 u_t^T\,dw_t^{\mathbb{Q}} - \frac{1}{2}\int_0^1 \|u_t\|^2\,dt
\end{align}
Substituting the above we get:
\begin{align}
    \ln p_\theta(x_{t_1}, \ldots, x_{t_n}) & \ge \mathbb{E}_{\mathbb{Q}}\!\left[\sum_{i=1}^n \ln p_\theta(x_{t_i} | z_{t_i}) -\ln \frac{q_0(z_0)}{p_0(z_0)} - \int_0^1 u_t^T\,dw_t^{\mathbb{Q}} - \frac{1}{2}\int_0^1 \|u_t\|^2\,dt\right] \\
    & = \mathbb{E}_{\mathbb{Q}}\!\left[\sum_{i=1}^n \ln p_\theta(x_{t_i} | z_{t_i}) -\ln \frac{q_0(z_0)}{p_0(z_0)} - \frac{1}{2}\int_0^1 \|u_t\|^2\,dt\right] \\
    & = \mathbb{E}_{\mathbb{Q}}\!\left[\sum_{i=1}^n \ln p_\theta(x_{t_i} | z_{t_i})\right]  - \mathrm{KL}(\mathbb{Q} \| \mathbb{P}_\theta)
\end{align}
Where the It\^o integral $\int_0^1 u_t^T\,dw_t^{\mathbb{Q}}$ vanishes under
$\mathbb{E}_{\mathbb{Q}}$ (as established in \Cref{thm:path_kl}).
\end{proof}

\begin{remark}
The bound has a natural interpretation: the first term is a
reconstruction likelihood (how well the model explains observations given the
latent path) and the second term penalizes the mismatch between the variational
and model path distributions.
$\frac{1}{2}\int_0^1 \|u_t\|^2$ can also be seen as the control cost required to
steer the model process $\mathbb{P}_\theta$ to match the variational process
$\mathbb{Q}$ \citep{theodorou2015nonlinear}.
The bound is tight when
$\mathbb{Q} = \mathbb{P}_\theta(\cdot \mid x_{t_1}, \ldots, x_{t_n})$,
i.e., when the variational process equals the true posterior process.
\end{remark}

\section{Observation-Space Stochastic Interpolants}
\label{sec:obs_space_si}

The LSI framework (\Cref{sec:lsi}) jointly trains an encoder, decoder,
and latent generative model. Here we consider the special case where the
generative process operates directly in observation space, without an encoder or
decoder. This corresponds to setting $z_t \equiv x_t$ for all $t$, making the
latent process identical to the observation process.

\paragraph{Setup:} The generative model is an SDE directly in observation space,
\begin{align}
\label{eq:obs_sde}
    d\tx_t = h_\theta(\tx_t, t)\,dt + \sigma_t\,d\tilde w_t, \quad \tx_0 \sim p_0(x_0)
\end{align}
i.e., \cref{eq:true_latent_dynamics} with $z \to x$. The prior $p_0(x_0)$
may be fixed or learnable (e.g., $p_\theta(x_0) = \mathcal{N}(\mu_\theta, \Sigma_\theta)$).
The marginal at $t=1$ defines the model distribution $p_\theta(x_1)$.

The variational process $\mathbb{Q}$ is constructed using the same diffusion
bridge machinery as in \Cref{sec:diffusion_bridge}, now bridging
$p_0(x_0)$ and $p_1(x_1) =p_{\text{data}}(x_1)$ directly in observation space.
As in the previous section, both $\mathbb{Q}$ and $\mathbb{P}_\theta$ can use
potentially different initial distributions $q_0$ and $p_0$.
Starting from the linear SDE 
$dx_t = h_t x_t\,dt + \sigma_t\,dw_t$ (cf.\ \cref{eq:gaussian_sde}) with
$z \to x$) and applying Doob's h-transform to condition on ending at
$x_1 \sim p_1$, the drift is (cf.\ \cref{eq:doobs_simple}):
\begin{align}
\label{eq:obs_bridge_drift}
    h_\phi(x_t, t) = h_t x_t + \sigma_t^2 \nabla_{x_t} \ln p(x_1 \mid x_t)
\end{align}
By construction, $\mathbb{Q}$ has marginal $p_{\text{data}} = p_1$ at $t=1$,
while $\mathbb{P}_\theta$ has marginal $p_\theta(x_1)$.

\paragraph{ELBO:} Since $p_{\text{data}} = p_1$ and $x_1$ is a
deterministic function of the path $\mathcal{X} \sim \mathbb{Q}$, the data processing inequality gives
\begin{align}
    \label{eq:likelihood_path_kl_bound}
    \mathrm{KL}(p_1 \| p_\theta) \le \mathrm{KL}(\mathbb{Q} \| \mathbb{P}_\theta)
\end{align}
Expanding the left side and rearranging yields the evidence lower bound:
\begin{align}
    \mathbb{E}_{p_1}[\ln p_\theta(x_1)] &\ge \mathbb{E}_{p_1}[\ln p_1(x_1)] - \mathrm{KL}(\mathbb{Q} \| \mathbb{P}_\theta) \label{eq:obs_elbo} \\
    &= -\mathrm{H}[p_1] - \mathrm{KL}(\mathbb{Q} \| \mathbb{P}_\theta) \label{eq:obs_elbo_entropy}
\end{align}
Where $\mathrm{H}[p_1]$ is the entropy of the data distribution $p_1$.
Using \Cref{thm:path_kl}, the path-space KL can be evaluated via
Girsanov's theorem as (with $z \to x$):
\begin{align}
\label{eq:obs_path_kl}
    \mathrm{KL}(\mathbb{Q} \| \mathbb{P}_\theta) = \mathrm{KL}(q_0 \| p_0) + \frac{1}{2}\,\mathbb{E}_{\mathbb{Q}}\!\left[\int_0^1 \|u(x_t, t)\|^2\,dt\right]
\end{align}
where $\sigma_t\,u(x_t, t) = h_\phi(x_t, t) - h_\theta(x_t, t)$, as in \cref{eq:u_def}.
If $q_0 = p_0$, the first term on the right vanishes, leaving only the dynamics
cost. As in main text, we assume $q_0 = p_0$ in the following.

\paragraph{Simulation-free training:} All the simulation-free machinery from
\cref{sec:lsi} carries over with $z \to x$. Using the observation-space
interpolant (cf.\ \cref{eq:lsi_interpolant}):
\begin{align}
\label{eq:obs_interpolant}
    x_t = \eta_t\,\epsilon + \kappa_t\,x_1 + \nu_t\,x_0, \quad \epsilon \sim \mathcal{N}(0, I)
\end{align}
$u(x_t, t)$ takes the following form -- similar to the result in \cref{sec:lsi} (cf.\ \cref{eq:general_u}):
\begin{align}
\label{eq:obs_u}
    u(x_t, t) = \sigma_t^{-1}\!\left[\left(\frac{d\eta_t}{dt} - \frac{\sigma_t^2}{2\eta_t}\right)\epsilon + \frac{d\kappa_t}{dt}\,x_1 + \frac{d\nu_t}{dt}\,x_0 - h_\theta(x_t, t)\right]
\end{align}

\paragraph{Final objective:} Substituting (\ref{eq:obs_path_kl}) and
(\ref{eq:obs_u}) into (\ref{eq:obs_elbo_entropy}), the observation-space ELBO is:
\begin{align}
\label{eq:obs_elbo_final}
    -\mathrm{H}[p_1] - \frac{1}{2}\,\mathbb{E}_{t \sim \mathcal{U}[0,1],\, x_1 \sim p_1,\, x_0 \sim p_0,\, \epsilon \sim \mathcal{N}(0,I)}\!\left[\|u(x_t, t)\|^2\right]
\end{align}
where the entropy $\mathrm{H}[p_1]$ of the data distribution is a constant and is
independent of the model parameters. For the linear choice
$\kappa_t = t$, $\nu_t = 1-t$, this specializes to the
following loss to be minimized (cf.\ \cref{eq:method_gen_elbo}):
\begin{align}
\label{eq:obs_elbo_linear}
    \frac{\beta_t}{2}\,\mathbb{E}\!\left[\left\|\sigma\sqrt{\frac{t}{1-t}}\,\epsilon + x_1 - x_0 - h_\theta(x_t, t)\right\|^2\right]
\end{align}
where $\beta_t$ has the same interpretation as in \cref{eq:method_gen_elbo}, of a
generalized weighting term, and the constant term has been dropped.

\begin{remark}
Comparing with the LSI loss (\cref{eq:method_gen_elbo}),
the observation-space ELBO is precisely the LSI objective with the
reconstruction term $-\ln p_\theta(x_1 \mid z_1)$ removed and $z \to x$
throughout. This confirms the consistency of the framework: LSI reduces to
observation-space stochastic interpolants when the encoder and decoder are
identity functions. All parameterizations (\Cref{sec:parameterization})
and sampling procedures (\Cref{sec:sampling}) apply directly with $z \to x$.
\end{remark}

\begin{remark}[Learnable prior]
The ELBO in \cref{eq:obs_elbo_final}
supports a learnable prior $p_\theta(x_0)$ without modification.
If both $\mathbb{Q}$ and $\mathbb{P}_\theta$ start from the same
$p_\theta(x_0)$, the initial KL vanishes regardless of the prior's form
(\Cref{thm:path_kl}). The prior parameters are still learned through the
interpolant $x_t = \eta_t \epsilon + \kappa_t x_1 + \nu_t x_0$ and the target
$\frac{d\nu_t}{dt} x_0$ in \cref{eq:obs_u}, providing gradients via the
reparameterization trick. Note that the interpolant coefficients
$\eta_t, \kappa_t, \nu_t$ depend only on the base SDE parameters
$h_t, \sigma_t$, not on $p_0$, so changing the prior affects only the sampling
distribution of $x_0$ --- not the interpolant structure.
\end{remark}

\section{Parameterizations}
\label{sec:parameterizations_app}

For the linear choice of $\kappa_t = t,\nu_t=1-t$ (\cref{sec:formulation_linear_app}) used for experiments in this paper, the loss term with $u(z_t, t)$ is
\begin{align}
\label{eq:general_u_loss}
\mathbb E_{t \sim \mathcal U[0, 1]}\mathbb E_{p(x_1, z_0, z_1)} \mathbb E_{p(z_t|z_1, z_0)} \frac{1}{2\sigma^2}\left\lVert-\sigma\sqrt{\frac{t}{1-t}}\epsilon + z_1 - z_0 - h_\theta(z_t, t)\right\rVert^2
\end{align}
Where $\epsilon \sim N(0, I)$. If $z_0$ is also Gaussian, $z_0\sim N(0, I)$, we can combine $\epsilon, z_0$ to yield $z_t = tz_1 + \sqrt{(1-t)(\sigma^2 t + 1 - t)}z_0$ and rewrite the above as
\begin{align}
\label{eq:general_u_loss_gauss}
    \mathbb E_{t \sim \mathcal U[0, 1]}\mathbb E_{p(x_1, z_0, z_1)} \mathbb E_{p(z_t|z_1, z_0)} \frac{1}{2}\left\lVert z_1-\sqrt{\frac{\sigma^2 t + 1 - t}{1-t}}z_0 - h_\theta(z_t, t)\right\rVert^2
\end{align}
Directly using above forms leads to high variance in gradients and unreliable training with frequent NaNs due to the $\sqrt{1-t}$ in the denominator. Consequently, we consider alternative parameterizations as discussed in the following. Two of the parameterizations \origflow{} and \flow{} are applicable for arbitrary $p_0$, while the remaining two \denoising{} and \noisepred{} are applicable when $z_0$ is Gaussian.
For each of these  parameterizations, we also derive the corresponding sampler in \cref{sec:detailed_sampling}

\subsection{\texorpdfstring{\origflow}{OrigFlow}}
With straightforward manipulation of the term inside the expectation we arrive at
\begin{align}
    & \frac{1}{2\sigma^2}\frac{1}{1-t}\left\lVert \sqrt{1-t}(z_1 - z_0) - \sigma \sqrt{t}\epsilon - \hat h_\theta(z_t, t)\right\rVert^2
\end{align}
where $\hat h_\theta(z_t, t) \equiv \sqrt{1-t}h_\theta(z_t, t)$. We rewrite above in terms of a time dependent weighting $\beta_t \equiv \frac{1}{\sigma^2(1-t)}$ as following.
\begin{align}
    & \frac{\beta_t}{2}\left\lVert \sqrt{1-t}(z_1 - z_0) - \sigma \sqrt{t}\epsilon - \hat h_\theta(z_t, t)\right\rVert^2
\end{align}
When $z_0$ is Gaussian, we can rewrite as
\begin{align}
    & \frac{\beta_t}{2}\left\lVert \sqrt{1-t}z_1 - \sqrt{\sigma^2t + 1 - t}z_0 - \hat h_\theta(z_t, t)\right\rVert^2
\end{align}
This objective can be viewed as estimating $\hat h_\theta(z_t, t) \equiv \mathbb E[\sqrt{1-t}z_1 - \sqrt{\sigma^2t + 1 - t}z_0 | z_t]$ with a time $t$ dependent weighting $\beta_t$.

\subsection{\texorpdfstring{\flow}{InterpFlow}}
Again, starting with the loss term with $u(z_t, t)$ and straightforward manipulations we arrive at the parameterization
\begin{align}
& \frac{1}{2\sigma^2}\left\lVert-\sigma\sqrt{\frac{t}{1-t}}\epsilon + z_1 - z_0 - h_\theta(z_t, t)\right\rVert^2 \\
&= \frac{1}{2\sigma^2}\left\lVert-\sigma\sqrt{\frac{t}{1-t}}\epsilon + z_1 - z_0 + \sqrt{\frac{t}{1-t}}z_t - \sqrt{\frac{t}{1-t}}z_t - h_\theta(z_t, t)\right\rVert^2 \\
&= \frac{\beta_t}{2}\left\lVert-\sigma\sqrt{t}\epsilon + \sqrt{1-t}(z_1 - z_0) + \sqrt{t}z_t - \hat h_\theta(z_t, t)\right\rVert^2
\end{align}
Where $\hat h_\theta(z_t, t) \equiv \sqrt{t}z_t + \sqrt{1-t}h_\theta(z_t, t)$ and $\beta_t \equiv \frac{1}{\sigma^2(1-t)}$. To gain insights into this parameterization, let's consider the term inside the norm and substitute $z_t$
\begin{align}
& -\sigma\sqrt{t}\epsilon + \sqrt{1-t}(z_1 - z_0) + \sqrt{t}z_t \\
&= -\sigma\sqrt{t}\epsilon + \sqrt{1-t}(z_1 - z_0) + \sqrt{t}(tz_1+(1-t)z_0+\sigma\sqrt{t(1-t)}\epsilon) \\
&= (\sqrt{1-t} + t\sqrt{t})z_1 + (\sqrt{t}(1-t)-\sqrt{1-t})z_0 + \sigma(t\sqrt{1-t}-\sqrt{t})\epsilon
\end{align}
Leading to
\begin{align}
& \frac{\beta_t}{2}\left\lVert (\sqrt{1-t} + t\sqrt{t})z_1 + (\sqrt{t}(1-t)-\sqrt{1-t})z_0 + \sigma(t\sqrt{1-t}-\sqrt{t})\epsilon  - \hat h_\theta(z_t, t)\right\rVert^2
\end{align}
The term $(\sqrt{1-t} + t\sqrt{t})z_1 + (\sqrt{t}(1-t)-\sqrt{1-t})z_0 + \sigma(t\sqrt{1-t}-\sqrt{t})\epsilon$ reduces to $z_1 - z_0$ at $t=0$ and $z_1 - \sigma\epsilon$ at $t=1$. Since this term appears to interpolate between the two, we refer to this parameterization as \flow{}.
When $z_0$ is also Gaussian, we can combine $\epsilon, z_0$ and rewrite as
\begin{align}
& \frac{\beta_t}{2}\left\lVert (\sqrt{1-t} + t\sqrt{t})z_1 + (\sqrt{t(1-t)}-1)\sqrt{\sigma^2t+1-t}z_0 - \hat h_\theta(z_t, t)\right\rVert^2
\label{eq:flow}
\end{align}
Observe that, with $\sigma = 1$, the term $(\sqrt{1-t} + t\sqrt{t})z_1 + (\sqrt{t(1-t)}-1)z_0$ reduces to $z_1 - z_0$ both at $t=0$ and $t=1$.

\subsection{\texorpdfstring{\denoising}{Denoising}}
This parameterization is applicable only when $z_0$ is Gaussian. Starting with the loss term with $u(z_t, t)$ and using the fact that $z_t = tz_1 + \sqrt{(1-t)(\sigma^2 t + 1 - t)}z_0$, we can manipulate the objective as following
\begin{align}
    &  \frac{1}{2}\left\lVert z_1-\sqrt{\frac{\sigma^2 t + 1 - t}{1-t}}z_0 - h_\theta(z_t, t)\right\rVert^2 \\
    &=  \frac{1}{2}\left\lVert z_1-\sqrt{\frac{\sigma^2 t + 1 - t}{1-t}}\frac{z_t - tz_1}{\sqrt{(1-t)(\sigma^2 t + 1 - t)}} - h_\theta(z_t, t)\right\rVert^2 \\
    &= \frac{1}{2}\left\lVert z_1 - \frac{z_t - tz_1}{1-t} - h_\theta(z_t, t)\right\rVert^2 \\
    &= \frac{1}{2}\frac{1}{(1-t)^2}\left\lVert z_1 - z_t - (1-t)h_\theta(z_t, t)\right\rVert^2 \\
    &= \frac{1}{2}\frac{1}{(1-t)^2}\left\lVert z_1 - \hat h_\theta(z_t, t)\right\rVert^2 \\
    &= \frac{\beta_t}{2}\left\lVert z_1 - \hat h_\theta(z_t, t)\right\rVert^2\label{eq:denoiser}
\end{align}
where  $\hat h_\theta(z_t, t) \equiv z_t + (1-t)h_\theta(z_t, t)$ and $\beta_t \equiv 1/ (1-t)^2$. In this form, $\hat h$ can be viewed as a denoiser. 

\subsection{\texorpdfstring{\noisepred}{NoisePred}}
This parameterization is applicable only when $z_0$ is Gaussian.
Similar to the previous section, we can construct the noise prediction parameterization by substituting $z_1$ using $z_t = tz_1 + \sqrt{(1-t)(\sigma^2 t + 1 - t)}z_0$.
\begin{align}
    &  \frac{1}{2}\left\lVert z_1-\sqrt{\frac{\sigma^2 t + 1 - t}{1-t}}z_0 - h_\theta(z_t, t)\right\rVert^2 \\
    &=  \frac{1}{2}\left\lVert \frac{z_t -  \sqrt{(1-t)(\sigma^2 t + 1 - t)}z_0}{t}-\sqrt{\frac{\sigma^2 t + 1 - t}{1-t}}z_0 - h_\theta(z_t, t)\right\rVert^2 \\
    &=  \frac{1}{2}\left\lVert \frac{\sqrt{1-t}z_t -  (1-t)\sqrt{\sigma^2 t + 1 - t}z_0-t\sqrt{\sigma^2 t + 1 - t}z_0}{t\sqrt{1-t}} - h_\theta(z_t, t)\right\rVert^2 \\
    &=  \frac{1}{2}\left\lVert \frac{\sqrt{1-t}z_t - \sqrt{\sigma^2 t + 1 - t}z_0}{t\sqrt{1-t}} - h_\theta(z_t, t)\right\rVert^2 \\
    &=  \frac{1}{2}\frac{1}{t^2(1-t)}\left\lVert \sqrt{1-t}z_t - \sqrt{\sigma^2 t + 1 - t}z_0 - t\sqrt{1-t}h_\theta(z_t, t)\right\rVert^2 \\
    &=  \frac{1}{2}\frac{\sigma^2 t + 1 - t}{t^2(1-t)}\left\lVert z_0 - \frac{\sqrt{1-t}z_t - t\sqrt{1-t}h_\theta(z_t, t)}{\sqrt{\sigma^2 t + 1 - t}}\right\rVert^2 \\
    &= \frac{\beta_t}{2}\left\lVert z_0 - \hat h_\theta(z_t, t)\right\rVert^2\label{eq:noise_predictor}
\end{align}
where  $\hat h_\theta(z_t, t) \equiv (\sqrt{1-t}z_t - t\sqrt{1-t}h_\theta(z_t, t))/\sqrt{\sigma^2 t + 1 - t}$ and $\beta_t \equiv 1/(t^2(1-t))$.

\section{Latent score function with Gaussian \texorpdfstring{$p_0$}{p0}}
\label{sec:score_fn}
When $p_0(z_0)$ is gaussian, $z_0 \sim N(0, I)$, we can compute the score function estimate $\nabla_{z_t}\ln p_t(z_t)$ from the learned drift $h_\theta$ \citep{singh2024stochastic}. When $z_0$ is gaussian, the transition density $p(z_t|z_1)$ is Gaussian.
With $z_t = \eta_t\epsilon + \kappa_t z_1 + \nu_t z_0$, we can reparameterize as $z_t = \kappa_t z_1 + \sqrt{\nu_t^2+\eta_t^2} z_0, z_0 \sim N(0, I)$.
\begin{align}
    p(z_t|z_1) = N(z_t; \kappa_t z_1, (\nu_t^2+\eta_t^2)I)
\end{align}
From \cite{singh2024stochastic}(eq. 41, Appendix B) we have
\begin{align}
    \nabla_{z_t} \ln p_t(z_t) = \mathbb E_{p_t(z_1 | z_t)}\left[ \frac{-z_t + \mu(z_1, t)}{\sigma(z_1, t)^2} \right]
\end{align}
Substituting
\begin{align}
    \nabla_{z_t} \ln p_t(z_t) &= \mathbb E_{p_t(z_1 | z_t)}\left[ \frac{-z_t + \kappa_t z_1}{\nu_t^2+\eta_t^2} \right] \\
    &= \frac{-z_t + \kappa_t \mathbb E[z_1|z_t]}{\nu_t^2+\eta_t^2} \label{eq:score_z1_zt} 
\end{align}
Since the interpolation relates $z_0, z_1, z_t$ as $z_t = \kappa_t z_1 + \sqrt{\nu_t^2+\eta_t^2} z_0$, we can rewrite the above expression in terms of $z_0$ as following
\begin{align}
  \nabla_{z_t} \ln p_t(z_t) &= -\frac{\E [z_0| z_t]}{\sqrt{\nu_t^2+\eta_t^2}}
\end{align}

\section{Latent score function with general \texorpdfstring{$p_0$}{p0}}
\label{sec:general_score_fn}
For a general distribution $p_0(z_0)$, it may not be possible to estimate the score function $\nabla_{z_t} \ln p_t(z_t)$ from the learned drift $h_\theta(z_t, t)$ alone. Here we derive the expression for estimating the score function for a general distribution $p_0(z_0)$. Recall from \cref{eq:gauss_zt_g_z0_z1} that $p(z_t | z_0, z_1)$ is Gaussian. From Denoising Score Matching \citep{vincent2011connection}, we can write
\begin{align}
\label{eq:dsm_solution}
\nabla_{z_t} \ln p_t(z_t) = \E_{p_t(z_0, z_1|z_t)} \frac{\partial \ln p_t(z_t | z_0, z_1)}{\partial z_t}
\end{align}
where we have conditioned on both variables $x_0, x_1$. Since $p(z_t | z_0, z_1)$ is Gaussian, as in the previous section, we can write
\begin{align}
    \nabla_{z_t} \ln p_t(z_t) = \mathbb E_{p_t(z_0, z_1 | z_t)}\left[ \frac{-z_t + \mu(z_0, z_1, t)}{\sigma(z_0, z_1, t)^2} \right]
\end{align}
Now, for $z_t = \eta_t\epsilon + \kappa_t z_1 + \nu_t z_0$, we have $p(z_t | z_0, z_1) = N(z_t; \kappa_t z_1 +\nu_t z_0, \eta_t^2I)$. Substituting
\begin{align}
    \nabla_{z_t} \ln p_t(z_t) &= \mathbb E_{p_t(z_0, z_1 | z_t)}\left[ \frac{-z_t + \kappa_t z_1 +\nu_t z_0}{\eta_t^2} \right] \\
    &= \mathbb E_{p_t(\epsilon | z_t)}\left[ \frac{-\eta_t\epsilon}{\eta_t^2} \right] \\
    &= -\frac{\mathbb E_{p_t(\epsilon | z_t)}[\epsilon]}{\eta_t} \equiv -\frac{\mathbb E[\epsilon | z_t]}{\eta_t}
\end{align}
Note that this result mirrors the one for SI(Theorem 2.8, \citep{albergo2023stochastic}),
though our derivation is straightforward and follows directly from Denoising
Score Matching \citep{vincent2011connection}.

\section{Detailed derivation of sampling}
\label{sec:detailed_sampling}
For an SDE of the form
\begin{align}
\label{eq:org_sde_rep1}
    dz_t = h_\theta(z_t, t)dt + \sigma_t dw_t
\end{align}
\cite{singh2024stochastic} (Corollary 1) derives a flexible family of samplers as following
\begin{align}
    \label{eq:general_sampler}
    dz_t &= \left[h_\theta(z_t, t) - \frac{(1-\gamma_t^2)\sigma_t^2}{2}\nabla_{z_t}\ln p_t(z_t)\right]dt + \gamma_t \sigma_t dw_t
\end{align}
where $\gamma_t$ is a time dependent weighting that can be chosen to control the amount of stochasticity injected into the sampling. Note that choosing $\gamma_t = 0$ yields the probability flow ODE \citep{song2020score} and results in a deterministic sampler. This general form of sampler requires both the drift $h_\theta(z_t, t)$ and the score function $\nabla_{z_t}\ln p_t(z_t)$. In general, the score function needs to be separately estimated. See \cref{sec:general_score_fn} for an estimator. We can also set $\gamma_t = 1$, leading to direct discretization of the original SDE in \cref{eq:org_sde_rep1}. 
However, for the special case of Gaussian $z_0$, we can infer the score function from the learned drift $h_\theta$ (\cref{sec:score_fn}). For this special case, we use the general form above to derive a family of samplers for various parameterizations discussed in \cref{sec:parameterizations_app}. Recall that for the choice of $\kappa_t = t, \nu_t=1-t$ used in this paper, the loss term is specified by \cref{eq:general_u_loss_gauss}. Without any reparameterization, we have
\begin{align}
    h_\theta(z_t, t) &= \frac{\mathbb E[z_1 | z_t] - z_t}{1 - t} \label{eq:htheta_denoiser}\\
    \mathbb E[z_1 | z_t] &= z_t + (1-t)h_\theta(z_t, t)
\end{align}
We can use the above to determine the expression for the score function
\begin{align}
    \nabla_x \ln p_t(z_t) &= \frac{-z_t + t h_\theta(z_t, t)}{\sigma^2t + 1 - t}
\end{align}
Above expressions for the score $\nabla_x \ln p_t(z_t)$ can then be plugged into \cref{eq:general_sampler} to derive a sampler for the original formulation
\begin{align}
  dz_t &= \left[h_\theta(z_t, t) - \frac{(1-\gamma_t^2)\sigma^2}{2}\frac{-z_t + t h_\theta(z_t, t)}{\sigma^2t + 1 - t} \right]dt + \gamma_t \sigma dw_t
\end{align}
For each of the following parameterizations, we calculate the expression for the drift $h_\theta$ and the score function $\nabla_x \ln p_t(z_t)$. These expressions can then be plugged into \cref{eq:general_sampler} to derive the sampler.

\subsection{Sampler for \texorpdfstring{\origflow}{OrigFlow}}
For the \origflow{} parameterization, we have
\begin{align}
    h_\theta(z_t, t) &= \frac{\hat h_\theta(z_t, t)}{\sqrt{1-t}}
\end{align}
For Gaussian $z_0$, we can now substitute into the expression for the score function
\begin{align}
    \nabla_x \ln p_t(z_t) &= \frac{-z_t + t h_\theta(z_t, t)}{\sigma^2t + 1 - t} \\
    &= \frac{-\sqrt{1-t}z_t + t \hat h_\theta(z_t, t)}{\sqrt{1-t}(\sigma^2t + 1 - t)}
\end{align}
The drift $h_\theta$ and the score function $\nabla_x \ln p_t(z_t)$ can now be plugged into \cref{eq:general_sampler} to derive the sampler.

\subsection{Sampler for \texorpdfstring{\flow}{InterpFlow}}
\label{sec:detailed_flow_sampler}
For the \flow{} parameterization, we have
\begin{align}
    h_\theta(z_t, t) &= \frac{\hat h(z_t, t) - \sqrt{t}z_t}{\sqrt{1 - t}}
\end{align}
For Gaussian $z_0$, we can now substitute into the expression for the score function
\begin{align}
    \nabla_x \ln p_t(z_t) &= \frac{-z_t + t h_\theta(z_t, t)}{\sigma^2t + 1 - t} \\
    &= \frac{-\sqrt{1-t}z_t + t \hat h_\theta(z_t, t) - t\sqrt{t}z_t}{\sqrt{1-t}(\sigma^2t + 1 - t)} \\
    &= \frac{-(\sqrt{1-t} + t\sqrt{t})z_t + t \hat h_\theta(z_t, t)}{\sqrt{1-t}(\sigma^2t + 1 - t)}
\end{align}
The drift $h_\theta$ and the score function $\nabla_x \ln p_t(z_t)$ can now be plugged into \cref{eq:general_sampler} to derive the sampler.

\subsection{Sampler for \texorpdfstring{\denoising}{Denoising}}
\label{sec:detailed_denoising_sampler}
For the \denoising{} parameterization, we have
\begin{align}
h_\theta(z_t, t) &= \frac{\hat h_\theta(z_t, t) - z_t}{1-t}
\end{align}
For Gaussian $z_0$, substituting into the expression for the score function
\begin{align}
    \nabla_x \ln p_t(z_t) &= \frac{-z_t + t h_\theta(z_t, t)}{\sigma^2t + 1 - t} \\
    &= \frac{-(1-t)z_t + t \hat h_\theta(z_t, t) - tz_t}{(1-t)(\sigma^2t + 1 - t)} \\
    &= \frac{-z_t + t \hat h_\theta(z_t, t)}{(1-t)(\sigma^2t + 1 - t)}
\end{align}
The drift $h_\theta$ and the score function $\nabla_x \ln p_t(z_t)$ can now be plugged into \cref{eq:general_sampler} to derive the sampler.

\subsection{Sampler for \texorpdfstring{\noisepred}{NoisePred}}
\label{sec:detailed_noisepred_sampler}
Again, we have
\begin{align}
h_\theta(z_t, t) &= \frac{-\sqrt{\sigma^2 t + 1 - t} \hat h_\theta(z_t, t) + \sqrt{1-t}z_t}{t\sqrt{1-t}}
\end{align}
For Gaussian $z_0$, substituting into the expression for the score function
\begin{align}
    \nabla_x \ln p_t(z_t) &= \frac{-z_t + t h_\theta(z_t, t)}{\sigma^2t + 1 - t} \\
    &= \frac{-\sqrt{1-t}z_t -  \sqrt{\sigma^2 t + 1 - t} \hat h_\theta(z_t, t) + \sqrt{1-t}z_t}{\sqrt{1-t}(\sigma^2t + 1 - t)} \\
    &= \frac{-\hat h_\theta(z_t, t)}{\sqrt{(1-t)(\sigma^2t + 1 - t)}} 
\end{align}
The drift $h_\theta$ and the score function $\nabla_x \ln p_t(z_t)$ can now be plugged into \cref{eq:general_sampler} to derive the sampler.

\section{Gaussianity of conditional density}
\label{sec:gaussian_conditional}
We have
\begin{align}
    p(z_t | z_1, z_0) &= \frac{p(z_1 | z_t)p(z_t | z_0)}{p(z_1 | z_0)}
\end{align}
Further, for the SDE  in \cref{eq:gaussian_sde}, using results from \cref{sec:transition_densities}, we have that the transition density $p(x_t | x_s)$ is normal with
\begin{align}
p(x_t|x_s) &= N(x_t; \mu_{st}, \Sigma_{st}) \\
\mu_{st} &= \mu_{s}\exp\left(\int_{s}^t h(\tau)d\tau\right) \equiv \mu_s a_{st}\\
\Sigma_{st} &= I\int_{s}^t \sigma(\tau)^2 \exp\left(2\int_\tau^t h(u)du\right)d\tau ) \equiv I b_{st}
\end{align}
Then, the conditional density $p(z_t | z_1, z_0)$ is also normal $N(z_t; \mu(z_0, z_1, t), \Sigma(z_0, z_1, t))$ with
\begin{align}
    \mu(z_0, z_1, t) &= \frac{b_{0t}a_{t1}z_1 + b_{t1}a_{0t}z_0}{b_{01}} \\
    \Sigma(z_0, z_1, t) &= \frac{b_{0t}b_{t1}}{b_{01}}I
\end{align}

\noindent
\textbf{Proof:}
First note that
\begin{align}
    a_{01} &= a_{0t}a_{t1} \\
    a_{st} &= \frac{a_{0t}}{a_{0s}} = \frac{a_{s1}}{a_{t1}} \\
    b_{st} &= \int_{s}^t \sigma(v)^2 a^2_{vt}dv
\end{align}
Next
\begin{align}
    b_{01} &= \int_{0}^1 \sigma(v)^2 a^2_{v1}dv \\
    &= \int_{0}^t \sigma(v)^2 a^2_{v1}dv + \int_{t}^1 \sigma(v)^2 a^2_{v1}dv \\
    &= \int_{0}^t \sigma(v)^2 a^2_{vt}a^2_{t1}dv + b_{t1} \\
    &= a^2_{t1}\int_{0}^t \sigma(v)^2 a^2_{vt}dv + b_{t1} \\
    &= a^2_{t1}b_{0t} + b_{t1} \label{eq:b01_identity}
\end{align}
Now
\begin{align}
p(z_t | z_1, z_0) &= \left({\frac{1}{2 \pi}\frac{b_{01}}{b_{t1}b_{0t}}}\right)^{\frac{n}{2}} \exp\left(-\frac{1}{2}\left(\frac{|z_1 - a_{t1}z_t|^2}{b_{t1}} + \frac{|z_t - a_{0t}z_0|^2}{b_{0t}} - \frac{|z_1 - a_{01}z_0|^2}{b_{01}}\right)\right)
\end{align}
Using the identities $a_{01} = a_{0t}a_{t1}, b_{01} = a^2_{t1}b_{0t} + b_{t1}$ and completing the squares we get
\begin{align}
p(z_t | z_1, z_0) &= \left({\frac{1}{2 \pi}\frac{b_{01}}{b_{0t}b_{t1}}}\right)^{\frac{n}{2}} \exp\left(-\frac{1}{2}\frac{b_{01}}{b_{0t}b_{t1}}\left|z_t - \frac{b_{0t}a_{t1}z_1 + b_{t1}a_{0t}z_0}{b_{01}}\right|^2\right)
\end{align}
We can therefore parameterize $z_t$ as following using the reparameterization trick.
\begin{align}
    z_t = \underbrace{\sqrt{\frac{b_{0t}b_{t1}}{b_{01}}}}_{\eta_t}\epsilon + \underbrace{\frac{b_{0t}a_{t1}}{b_{01}}}_{\kappa_t}z_1 + \underbrace{\frac{b_{t1}a_{0t}}{b_{01}}}_{\nu_t}z_0, \quad \epsilon \sim N(0, I)
\end{align}
we can succinctly rewrite the above as
\begin{align}
    z_t = \eta_t\epsilon + \kappa_t z_1 + \nu_t z_0, \quad \epsilon \sim N(0, I)
\end{align}
Where $\eta_t, \kappa_t, \nu_t$ are appropriate scalar functions of time $t$.

%%%%%%%%%%%%%%%%%%%%%%%%%%%%%%%%%%%%%%%%%%%%
\section{General training objective}
\label{sec:general_training_obj_app}
Here we derive the form of the general training objective. The first term in the objective  is the reconstruction term and remains as is. The second term of the training objective uses $u(z_t, t)$, let's recall it's expression
\begin{align}
u(z_t, t) &= \sigma_t^{-1}[h_tz_t + \sigma^2_t\nabla_{z_t}\ln p(z_1|z_t) - h_\theta(z_t, t)]
\end{align}
The first two terms in the above serve as the target for $h_\theta$. Next, we rewrite them in terms of existing variables. Let $\xi(t)$ denote these two terms and substitute \cref{eq:grad_log_z1_zt} as following
\begin{align}
    \xi(t) &= h_tz_t + \sigma^2_t\nabla_{z_t}\ln p(z_1|z_t) \\
    &= h_t z_t + \frac{\sigma_t^2 a_{t1}(z_1 - a_{t1}z_t)}{b_{t1}} \\
    &= \left(h_t - \frac{\sigma_t^2a_{t1}^2}{b_{t1}}\right)z_t + \frac{\sigma_t^2 a_{t1}z_1}{b_{t1}} \label{eq:xi_init_form}
\end{align}
Next, recall the stochastic interpolant and the expressions for $a_{st}$ and $b_{st}$ from \cref{sec:gaussian_conditional}
\begin{align}
    z_t &= \eta_t\epsilon + \kappa_t z_1 + \nu_t z_0, \quad \epsilon \sim N(0, I) \\
    \eta_t &= \sqrt{\frac{b_{0t}b_{t1}}{b_{01}}}, \quad
    \kappa_t = \frac{b_{0t}a_{t1}}{b_{01}}, \quad
    \nu_t = \frac{b_{t1}a_{0t}}{b_{01}}, \\
    a_{st} &= \exp\left(\int_s^th(\tau)d\tau\right), \quad
    b_{st} = \int_{s}^t \sigma(v)^2 a^2_{vt}dv \\
\end{align}
Intuitively, we expect the drift $h_\theta$ to be related to the velocity field. Therefore, we compute the time derivatives of $\kappa_t, \nu_t$ and $\eta_t$ next
\begin{align}
    \frac{d\kappa_t}{dt} &= \frac{1}{b_{01}}\left(b_{0t}\frac{d a_{t1}}{dt} + \frac{d b_{0t}}{dt}a_{t1}\right) \\
    \frac{d\nu_t}{dt} &= \frac{1}{b_{01}}\left(b_{t1}\frac{d a_{0t}}{dt} + \frac{d b_{t1}}{dt}a_{0t}\right) \\
    \frac{d\eta_t}{dt} &= \frac{1}{2\eta_t b_{01}}\left(b_{0t}\frac{d b_{t1}}{dt} + \frac{d b_{0t}}{dt}b_{t1}\right)
\end{align}
From the expression for $a_{st}$, using differentiation under the integral sign, we have
\begin{align}
    \frac{d a_{0t}}{dt} &= a_{0t}h_t, \quad \frac{d a_{t1}}{dt} = -a_{t1}h_t
\end{align}
Similarly, from the expression for $b_{st}$
\begin{align}
    \frac{d b_{0t}}{dt} &= \sigma_t^2 a_{tt}^2 + 2\int_0^t \sigma(v)^2 a_{vt}^2h_t dv = \sigma_t^2 + 2b_{0t}h_t \\
    \frac{d b_{t1}}{dt} &= -\sigma_t^2 a_{t1}^2
\end{align}
Since $a_{tt} = 1$. Substituting back into the equations for the derivatives of $\kappa_t$  and $\nu_t$
\begin{align}
    \frac{d\kappa_t}{dt} &= \frac{1}{b_{01}}\left(-b_{0t}a_{t1}h_t + (\sigma_t^2 + 2b_{0t}h_t)a_{t1}\right)
    = \frac{1}{b_{01}}\left(\sigma_t^2a_{t1} + b_{0t}a_{t1}h_t\right) \\
    &= \frac{\sigma_t^2a_{t1}}{b_{01}} + \kappa_th_t \\
    \frac{d\nu_t}{dt} &= \frac{1}{b_{01}}\left(b_{t1}a_{0t}h_t -\sigma_t^2 a_{t1}^2a_{0t}\right)
    = \nu_t h_t -\frac{\sigma_t^2 a_{t1}^2a_{0t}}{b_{01}} \\
    &= \nu_t \left(h_t -\frac{\sigma_t^2 a_{t1}^2}{b_{t1}}\right) \\
    \frac{d\eta_t}{dt} &= \frac{1}{2\eta_t b_{01}}\left(-\sigma_t^2 a_{t1}^2b_{0t} + (\sigma_t^2 + 2b_{0t}h_t)b_{t1}\right) \\
    &= \frac{1}{2\eta_t b_{01}}\left((b_{t1} - a_{t1}^2b_{0t})\sigma_t^2 +  2b_{0t}b_{t1}h_t\right) \\
    &= \frac{1}{2\eta_t b_{01}}\left((b_{t1} - a_{t1}^2b_{0t})\sigma_t^2 +  2b_{0t}\left( \frac{b_{t1}}{\nu_t}\frac{d\nu_t}{dt}+\sigma_t^2a_{t1}^2\right)\right) \\
    &= \frac{1}{2\eta_t b_{01}}\left((b_{t1} + a_{t1}^2b_{0t})\sigma_t^2 +  \frac{2b_{0t}b_{t1}}{\nu_t}\frac{d\nu_t}{dt}\right) \\
    &= \frac{1}{2\eta_t b_{01}}\left(b_{01}\sigma_t^2 +  \frac{2\eta_t^2b_{01}}{\nu_t}\frac{d\nu_t}{dt}\right) \\
    &= \frac{\sigma_t^2}{2\eta_t } +  \frac{\eta_t}{\nu_t}\frac{d\nu_t}{dt}
\end{align}
Where we have used the identity $b_{01} = b_{t1} + a_{t1}^2b_{0t}$ from \cref{eq:b01_identity}. Further, we can relate $\frac{d\kappa_t}{dt}$ and $\frac{d\nu_t}{dt}$ by eliminating $h_t$ as following
\begin{align}
    \frac{d\kappa_t}{dt} &= \frac{\sigma_t^2a_{t1}}{b_{01}} + \kappa_t \left(\frac{1}{\nu_t}\frac{d\nu_t}{dt} + \frac{\sigma_t^2 a_{t1}^2}{b_{t1}}\right) 
    = \frac{\kappa_t}{\nu_t}\frac{d\nu_t}{dt} +\frac{\sigma_t^2a_{t1}}{b_{01}}+ \kappa_t\frac{\sigma_t^2 a_{t1}^2}{b_{t1}} \\
    &= \frac{\kappa_t}{\nu_t}\frac{d\nu_t}{dt} +\frac{\sigma_t^2a_{t1}}{b_{01}}+ \frac{b_{0t}a_{t1}}{b_{01}}\frac{\sigma_t^2 a_{t1}^2}{b_{t1}}
    = \frac{\kappa_t}{\nu_t}\frac{d\nu_t}{dt} + \frac{\sigma_t^2a_{t1}(b_{t1} + b_{0t}a_{t1}^2)}{b_{01}b_{t1}} \\
    &= \frac{\kappa_t}{\nu_t}\frac{d\nu_t}{dt} + \frac{\sigma_t^2a_{t1}b_{01}}{b_{01}b_{t1}} \\
    &= \frac{\kappa_t}{\nu_t}\frac{d\nu_t}{dt} + \frac{\sigma_t^2a_{t1}}{b_{t1}}
\end{align}
We can now substitute into the expression for $\xi(t)$ in \cref{eq:xi_init_form}
\begin{align}
    \xi(t) &= \frac{1}{\nu_t}\frac{d\nu_t}{dt}z_t + \frac{\sigma_t^2 a_{t1}z_1}{b_{t1}} \\
    &= \frac{1}{\nu_t}\frac{d\nu_t}{dt}(\eta_t\epsilon + \kappa_t z_1 + \nu_t z_0) + \left(\frac{d\kappa_t}{dt}-\frac{\kappa_t}{\nu_t}\frac{d\nu_t}{dt}\right)z_1 \\
    &= \frac{\eta_t}{\nu_t}\frac{d\nu_t}{dt}\epsilon + \frac{d\kappa_t}{dt}z_1 + \frac{d\nu_t}{dt}z_0 \\
    &= \left(\frac{d\eta_t}{dt}-\frac{\sigma_t^2}{2\eta_t }\right)\epsilon + \frac{d\kappa_t}{dt}z_1 + \frac{d\nu_t}{dt}z_0
\end{align}
Substituting back into the expression for $u(z_t, t)$ we can write the general form as following
\begin{align}
u(z_t, t) &= \sigma_t^{-1}\left[\left(\frac{d\eta_t}{dt}-\frac{\sigma_t^2}{2\eta_t }\right)\epsilon + \frac{d\kappa_t}{dt}z_1 + \frac{d\nu_t}{dt}z_0 - h_\theta(z_t, t)\right]
\end{align}
With the $u(z_t, t)$ above, the ELBO can be written using \cref{eq:sde_elbo}.

%%%%%%%%%%%%%%%%%%%%%%%%%%%%%%%%%%%%%%%%%%%%
\section{Drift \texorpdfstring{$h_t$}{ht}, dispersion \texorpdfstring{$\sigma_t$}{sigmat} and stochasticity \texorpdfstring{$\eta_t$}{etat}  from \texorpdfstring{$\kappa_t, \nu_t$}{kt, vt}}
\label{sec:general_drift_dispersion}
Often, specifying the interpolant coefficients $\kappa_t, \nu_t$ is intuitively easier than specifying $h_t, \sigma_t$ directly. Here we derive expressions for $h_t$ and $\sigma_t$ given $\kappa_t$ and $\nu_t$. We have
\begin{align}
    \frac{d\kappa_t}{dt} &= \kappa_th_t + \frac{\sigma_t^2a_{t1}}{b_{01}} \label{eq:dd_base_1_app}\\
    \frac{d\nu_t}{dt} &= h_t\nu_t -\frac{\sigma_t^2 a_{t1}^2}{b_{t1}}\nu_t \label{eq:dd_base_2_app}
\end{align}
Multiplying first equation by $\nu_t$ and second by $\kappa_t$ and then subtracting the second from the first
\begin{align}
    \nu_t\frac{d\kappa_t}{dt}-\kappa_t\frac{d\nu_t}{dt} &= \nu_t\frac{\sigma_t^2a_{t1}}{b_{01}} + \kappa_t\frac{\sigma_t^2 a_{t1}^2}{b_{t1}}\nu_t \\
    &= \left(\nu_t\frac{\sigma_t^2a_{t1}}{b_{01}} + \kappa_t\frac{\sigma_t^2 a_{t1}^2}{b_{t1}}\nu_t \right)
\end{align}
Substituting in the definitions of $\kappa_t$ and $\nu_t$ in RHS and simplifying
\begin{align}
    \nu_t\frac{d\kappa_t}{dt}-\kappa_t\frac{d\nu_t}{dt} &= \left(\frac{b_{t1}a_{01}\sigma_t^2}{b_{01}^2} + \frac{b_{0t}\sigma_t^2 a_{t1}^2 a_{01}}{b_{01}^2} \right) \\
    &= \frac{a_{01}\sigma_t^2}{b_{01}^2}\left(b_{t1} + b_{0t}a_{t1}^2 \right) 
    = \frac{a_{01}\sigma_t^2}{b_{01}^2}b_{01} \\
    &= \frac{a_{01}\sigma_t^2}{b_{01}}
\end{align}
where we have used $a_{01}=a_{0t}a_{t1}$ and $b_{01} = b_{t1} + b_{0t}a_{t1}^2$.
Therefore
\begin{align}
    \sigma_t^2 = \frac{b_{01}}{a_{01}}\left(\nu_t\frac{d\kappa_t}{dt}-\kappa_t\frac{d\nu_t}{dt} \right) \label{eq:sigma2_t_expression_app}
\end{align}
Where $b_{01} > 0, a_{01} > 0$ are time $t$ independent constants that can't be determined by $\kappa_t, \nu_t$ alone. In this paper, we assume $a_{01}=2$ and $b_{01}=a_{01}\sigma^2$, where $\sigma$ is a hyper-parameter. Next, to derive the expression for $h_t$, we eliminate $\sigma_t^2$ from \cref{eq:dd_base_1_app,eq:dd_base_2_app}.
\begin{align}
    b_{01}\left(\frac{d\kappa_t}{dt} - \kappa_th_t\right) &=  
    \frac{b_{t1}}{a_{t1}}\left(-\frac{1}{\nu_t}\frac{d\nu_t}{dt} + h_t\right) \\
    h_t\left(b_{01}\kappa_t + \frac{b_{t1}}{a_{t1}}\right) &= b_{01}\frac{d\kappa_t}{dt} +\frac{b_{t1}}{a_{t1}}\frac{1}{\nu_t}\frac{d\nu_t}{dt} \\
    h_t\left(\frac{a_{t1}b_{01}\kappa_t + b_{t1}}{a_{t1}}\right) &= b_{01}\frac{d\kappa_t}{dt} +\frac{b_{t1}}{a_{t1}}\frac{b_{01}}{b_{t1}a_{0t}}\frac{d\nu_t}{dt} \\
    h_t\left(a_{0t}a_{t1}\kappa_t + \frac{a_{0t}b_{t1}}{b_{01}}\right) &= a_{0t}a_{t1}\frac{d\kappa_t}{dt} + \frac{d\nu_t}{dt} \\
    h_t\left(a_{01}\kappa_t + \nu_t\right) &= a_{01}\frac{d\kappa_t}{dt} + \frac{d\nu_t}{dt} \\
    h_t&= \frac{a_{01}\frac{d\kappa_t}{dt} + \frac{d\nu_t}{dt}}{a_{01}\kappa_t + \nu_t} \label{eq:h_t_expression_app}
\end{align}
As before, $a_{01} > 0$ is a time independent constant that can't be determined from the choice of $\kappa_t, \nu_t$ alone. Finally, to express $\eta_t$ in terms of given $\kappa_t, \nu_t$, note that
\begin{align}
    \eta_t^2 &= \frac{b_{0t}b_{t1}}{b_{01}} = \frac{b_{01}}{a_{0t}a_{t1}}\frac{b_{0t}a_{t1}}{b_{01}} \frac{b_{t1}a_{0t}}{b_{01}} = \frac{b_{01}}{a_{01}} \kappa_t \nu_t
\end{align}
where we have used the identity $a_{01}=a_{0t}a_{t1}$. In the following, we derive the formulation for the linear $\kappa_t, \nu_t$ schedule used in experiments in this paper. This schedule also corresponds to the choice used in Stochastic Interpolants\citep{albergo2023stochastic}. Note that similar choice is made by the Rectified Flow \citep{liu2022flow}, however the missing $\eta$ term implies that they do not have a bound on the likelihood, as also observed by \cite{albergo2023stochastic}. We also provide the derivation for the variance preserving schedule as it is quite commonly used for diffusion models. However, we do not empirically explore it.

\section{Formulation for linear \texorpdfstring{$\kappa_t, \nu_t$}{kt, vt}}
\label{sec:formulation_linear_app}
For linear choice $\kappa_t=t, \nu_t=1-t$. Further, we assume $a_{01}=2, b_{01}=a_{01}\sigma^2 $. Therefore,
\begin{align}
    \frac{d\kappa_t}{dt} &= 1, \quad \frac{d\nu_t}{dt} = -1
\end{align}
We can write the expressions for $h_t$ and $\sigma_t^2$ directly, using \cref{eq:h_t_expression_app,eq:sigma2_t_expression_app}, as
\begin{align}
    h_t &= \frac{1}{1 + t}, \quad \sigma_t^2 = \sigma^2
\end{align}
To express the latent stochastic interpolant, we can calculate the coefficient $\eta_t$ for $\epsilon$
\begin{align}
    \eta_t = \sqrt{\frac{b_{01}}{a_{01}} \kappa_t \nu_t} = \sigma \sqrt{t(1-t)}
\end{align}
We can now write the expression for the latent stochastic interpolant
\begin{align}
    z_t &= \sigma \sqrt{t(1-t)}\epsilon + tz_1 + (1-t)z_0, \quad \epsilon \sim N(0, I).
\end{align}
Finally, to express $u(z_t, t)$ first we calculate
\begin{align}
    \frac{d\eta_t}{dt}-\frac{\sigma_t^2}{2\eta_t } &= \frac{\sigma(1-t-t)}{2\sqrt{t(1-t)}} - \frac{\sigma^2}{2\sigma \sqrt{t(1-t)}} = \frac{\sigma^2(1-2t)-\sigma^2}{2\sigma \sqrt{t(1-t)}} = -\sigma\sqrt{\frac{t}{1-t}}
\end{align}
leading to
\begin{align}
u(z_t, t) &= \sigma^{-1}\left[-\sigma\sqrt{\frac{t}{1-t}}\epsilon + z_1 - z_0 - h_\theta(z_t, t)\right]
\end{align}

\section{Formulation for variance preserving \texorpdfstring{$\kappa_t, \nu_t$}{kt, vt}}
\label{sec:formulation_var_preserve_app}

For the variance preserving formulation, we set $\kappa_t = \sqrt{t}$ and $\eta_t^2 + \nu_t^2 = 1 - t$. Note that if $z_0 \sim N(0, I)$ is Gaussian, this setting leads to the latent stochastic interpolant $z_t = \sqrt{t}z_1 + \sqrt{1-t}z_0$. Here $\epsilon$ and $z_0$ have been combined since they both are Gaussian. Let $b_{01}/a_{01}=C$, then
\begin{align}
    \eta_t^2 &= C\sqrt{t} \nu_t = 1 - t - \nu_t^2 \\
    \implies \nu_t &= \frac{-C\sqrt{t} + \sqrt{(C^2 - 4)t + 4}}{2}
\end{align}
Using above, the expressions for $h_t$ and $\sigma_t^2$ can be derived as
\begin{align}
    h_t &= \frac{\frac{a_{01}}{\sqrt{t}} - \frac{C}{2\sqrt{t}} + \frac{C^2-4}{2\sqrt{(C^2-4)t+4}}}{2a_{01}\sqrt{t}-C\sqrt{t}+\sqrt{(C^2-4)t+4}} \\
    \sigma_t^2 &= \frac{C}{\sqrt{t}\sqrt{(C^2-4)t+4}}
\end{align}
Choosing $a_{01}=1$ and $C=2$ yields
\begin{align}
    h_t &= 0, \quad 
    \sigma_t^2 = \frac{1}{\sqrt{t}}, \quad
    \nu_t = 1 - \sqrt{t}
\end{align}
The coefficient $\eta_t$ for $\epsilon$ can be calculated as
\begin{align}
    \eta_t = \sqrt{\frac{b_{01}}{a_{01}} \kappa_t \nu_t} = \sqrt{2 \sqrt{t}(1 - \sqrt{t})}
\end{align}
We can now write the expression for the latent stochastic interpolant
\begin{align}
    z_t &= \sqrt{2 \sqrt{t}(1 - \sqrt{t})}\epsilon + \sqrt{t}z_1 + (1-\sqrt{t})z_0, \quad \epsilon \sim N(0, I).
\end{align}
Finally, to express $u(z_t, t)$ first we calculate
\begin{align}
    \frac{d\eta_t}{dt}-\frac{\sigma_t^2}{2\eta_t } &= -\frac{1}{\sqrt{2 \sqrt{t}(1 - \sqrt{t})}}
\end{align}
with
\begin{align}
    \frac{d\kappa_t}{dt} = \frac{1}{2\sqrt{t}}, \quad \frac{d\nu_t}{dt} = -\frac{1}{2\sqrt{t}}
\end{align}
we arrive at
\begin{align}
u(z_t, t) &= \sigma^{-1}\left[-\frac{1}{\sqrt{2 \sqrt{t}(1 - \sqrt{t})}}\epsilon + \frac{1}{2\sqrt{t}}z_1 - \frac{1}{2\sqrt{t}}z_0 - h_\theta(z_t, t)\right]
\end{align}
Note that above expression is for a particular choice of $a_{01}=1$ and the ratio $b_{01}/a_{01}=2$, which we chose for relative simplicity of the final expression above. Other choices can be made, leading to different expressions.

%%%%%%%%%%%%%%%%%%%%%%%%%%%%%%%%%%%%%%%%%%%%
\section{Gaussian transition densities}
\label{sec:transition_densities}

Let's consider a linear SDE of the form 
\begin{align}
    dz_t = h_tz_tdt + u_tdt + \sigma_t dw_t
\end{align}
When the SDE is linear with additive noise, we know that the transition densities are gaussian and are therefore fully specified by their mean and covariance. From ~\cite{sarkka2019applied} (Eq 6.2) these are specified by the following differential equations
\begin{align}
    \frac{d\mu_t}{dt} &= h_t\mu_t + u_t\\
    \frac{d\Sigma_t}{dt} &= 2h_t\Sigma_t + \sigma_t^2I
\end{align}
The solution to these is given by (eq. 6.3, 6.4, \cite{sarkka2019applied})
\begin{align}
    \mu_t &= \Psi(t, t_0)\mu_{t_0} + \int_{t_0}^t \Psi(t, \tau)u(\tau)d\tau \\
    \Sigma_t &= \Psi(t, t_0)\Sigma_{t_0} \Psi(t, t_0)^T + \int_{t_0}^t \sigma(\tau)^2\Psi(t, \tau)\Psi(t, \tau)^Td\tau
\end{align}
Where $\Psi(s, t)$ is the transition matrix. For our specific case of linear SDEs, we have
\begin{align}
    \Psi(s, t) = \exp\left(\int_t^s h(\tau)d\tau\right)
\end{align}
Substituting, we get
\begin{align}
    \mu_t &= \mu_{t_0}\exp\left(\int_{t_0}^t h(\tau)d\tau\right) + \int_{t_0}^t \exp\left(\int_\tau^t h(s)ds\right)u(\tau)d\tau \\
    \Sigma_t &= \Sigma_{t_0}\exp\left(2\int_{t_0}^t h(\tau)d\tau\right) + I\int_{t_0}^t \sigma(\tau)^2 \exp\left(2\int_\tau^t h(s)ds\right)d\tau
\end{align}

\section{Gaussian \texorpdfstring{$z_0$}{z0}}
\label{sec:gaussian_z_0}
\label{sec:simplified_u}

For the interpolant (\cref{sec:formulation_linear_app})
\begin{align}
\label{eq:z_t_interp_gen}
    z_t &= \sigma \sqrt{t(1-t)}\epsilon + tz_1 + (1-t)z_0, \quad \epsilon \sim N(0, I),
\end{align}
if $z_0$ is gaussian, we can replace the linear combination of two normal random variables $\epsilon, z_0$ with a single random variable $\hat z_0 \sim N(\hat \mu, \hat \Sigma)$. Assuming $z_0 \sim N(0, I)$, the mean $\hat \mu = 0$ and covariance $\hat \Sigma$ can be computed as 
\begin{align}
    \hat \Sigma &= \left(\sigma^2t(1-t) + (1-t)^2\right)I \\
    &= (1-t)(t\sigma^2 + (1-t))I
\end{align}
Using the reparameterization trick, we can express $\hat z_0$ in terms of $z_0$ and write
\begin{align}
    z_t = tz_1 + \sqrt{(1-t)(t\sigma^2 + (1-t))}z_0, \quad z_0 \sim N(0, I)
\end{align}
Note that
\begin{align}
    z_t &= tz_1 + \sqrt{1-t}z_0, && \quad \text{if}~\sigma^2 = 1 \\
    z_t &= tz_1 + (1-t) z_0, &&\quad \text{if}~\sigma^2 = 0
\end{align}
Similarly, recall the expression for $u(z_t, t)$ from \cref{sec:formulation_linear_app}
\begin{align}
u(z_t, t) &=\sigma^{-1}\left[-\sigma\sqrt{\frac{t}{1-t}}\epsilon + z_1 - z_0 - h_\theta(z_t, t)\right]
\end{align}
If $z_0 \sim N(0, I)$ is also gaussian, we can combine $\epsilon, z_0$ and write
\begin{align}
u(z_t, t)&= \sigma^{-1}\left[z_1-\sqrt{\frac{1 + (\sigma^2-1)t}{1-t}}z_0 - h_\theta(z_t, t)\right]
\end{align}
if we choose $\sigma^2=1$, then the expression simplifies to
\begin{align}
u(z_t, t) &= z_1 - \frac{1}{\sqrt{1-t}}z_0 - h_\theta(z_t, t)
\end{align}

Finally, we would like to reiterate that we arrive at the above by assuming $z_0$ is gaussian. The general form derived in other sections make no assumptions about the distribution of $z_0$.

\begin{figure}[t]
  \centering
  \begin{tikzpicture}
    \begin{axis}[
        xlabel={$s$},          % Label for the x-axis
        ylabel={$t(s)$},       % Label for the y-axis
        xmin=0, xmax=1,        % Range for the x-axis
        ymin=0, ymax=1.05,     % Range for the y-axis (slightly above 1 for legend)
        legend pos=south east, % Position of the legend
        legend style={cells={anchor=west}}, % Align legend entries to the left
        grid=major,            % Add a major grid
        % smooth,              % 'smooth' can be added if desired.
                               % Plotting from discrete points might not need it.
        title={$t(s) = 1 - (1-s)^c$}, % Title
        height=8cm,            % Height of the plot
        width=10cm,            % Width of the plot
    ]

    % Ensure 'ts_plot_data.csv' is in the 'figs' subdirectory
    % relative to your .tex file.
    % The column names in the CSV are 's', 'c0p1', 'c0p2', 'c0p5', 'c1', 'c2', 'c5', 'c10'
    % as generated by the Python script with id="python_csv_generator_for_pgfplots"

    % Plot for c = 0.1
    \addplot[color=teal, thick] table [x=s, y=c0p1, col sep=comma] {figs/ts_plot_data.csv};
    \addlegendentry{$c=0.1$}

    % Plot for c = 0.2
    \addplot[color=violet, thick] table [x=s, y=c0p2, col sep=comma] {figs/ts_plot_data.csv};
    \addlegendentry{$c=0.2$}

    % Plot for c = 0.5
    \addplot[color=lime, thick] table [x=s, y=c0p5, col sep=comma] {figs/ts_plot_data.csv};
    \addlegendentry{$c=0.5$}

    % Plot for c = 1
    \addplot[color=blue, thick] table [x=s, y=c1, col sep=comma] {figs/ts_plot_data.csv};
    \addlegendentry{$c=1$}

    % Plot for c = 2
    \addplot[color=red, thick] table [x=s, y=c2, col sep=comma] {figs/ts_plot_data.csv};
    \addlegendentry{$c=2$}

    % Plot for c = 5
    \addplot[color=green!60!black, thick] table [x=s, y=c5, col sep=comma] {figs/ts_plot_data.csv};
    \addlegendentry{$c=5$}

    % Plot for c = 10
    \addplot[color=orange, thick] table [x=s, y=c10, col sep=comma] {figs/ts_plot_data.csv};
    \addlegendentry{$c=10$}

    \end{axis}
\end{tikzpicture}
\caption{\textbf{Schedule for $t$.} A visualization of the schedule for $t(s)$ with $s \in [0, 1]$ as $c$ is varied. As $c$ increases, larger $t$ values are favored, thereby sampling interpolants closer to $t=1$ more frequently.}.
 \label{fig:t_schedule}
\end{figure}
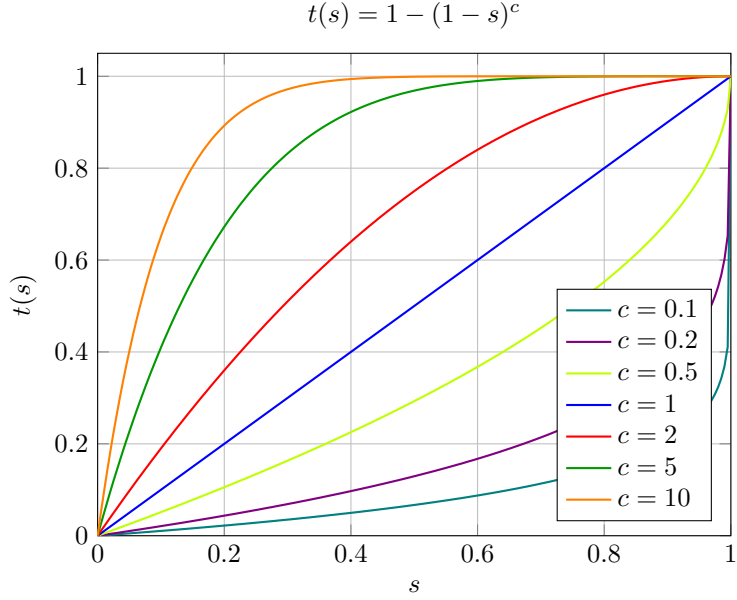

%%%%%%%%%%%%%%%%%%%%%%%%%%%%%%%%%%%%%%%%%%%%%%%%%%%%%%%%%%%%%%%%%%%%%
\section{Choice of prior \texorpdfstring{$p_0$}{p0}}
\label{sec:choice_of_priors}

The Gaussian distribution, along with a small set of other distributions, enjoys the special privilege of being Lévy stable. That is, a linear combination of two Gaussian random variables is still a Gaussian random variable. Lévy stability is the main property behind the original formulation of the simulation free training of the Gaussian diffusion models, e.g. as in DDPM.
In contrast, Laplacian, Uniform and Gaussian Mixture are not Lévy stable, and thus our experiment with those provides strong evidence for the general nature of the proposed method.
The Gaussian mixture used in our experiment was constructed by having a component for each training image. Consequently, it is a mixture with a very large number of components.
The current estimate of the encoder being learned was used to encode the training images, yielding the means of the corresponding components. Standard deviation for each dimension was fixed to $0.1$.
In practice, we simply shuffled the encoding of the training images, added noise, and used a $\mathtt{stop\_gradient}$ operation to prevent the flow of gradient through the prior. Since the encoder is also evolving during training, this experiment required $\sim 3\times$ more steps to yield the reported FID. Without $\mathtt{stop\_gradient}$, the experiment became unstable.

%%%%%%%%%%%%%%%%%%%%%%%%%%%%%%%%%%%%%%%%%%%%%%%%%%%%%%%%%%%%%%%%%%%%%
\section{ImageNet training and evaluation details}
\label{sec:add_imagenet_details}

We trained our models using the entire ImageNet training dataset, consisting of approximately 1.2 million images.
Models are trained with Stochastic Gradient Descent (SGD) with the AdamW optimizer \citep{kingma2014adam,loshchilov2017decoupled}, using $\beta_1=0.9, \beta_2=0.99, \epsilon=10^{-12}$. All models are trained for $1000$ epochs using a batch size of $2048$, except for the ones reported in \cref{tab:resolution_fid} where they were trained for 2000 epochs. Only center crops were used after resizing the images to the have the smaller side match the target resolution. For data augmentation, only horizontal (left-right) flips were used. Pixel values for an image $I$ were scaled to the range $[-1, 1]$ by computing $2(I / 255)-1$ before feeding to the model. For evaluation, a exponential moving average of the model's parameters was used using a decay rate of $0.9999$. The FIDs were computed over the training dataset, with reference statistics derived from center-cropped images, without any further augmentation. All FIDs are reported with class conditioned samples. To compute PSNR, sampled image pixel values were scaled back to the range $[0, 255]$ and quantized to integer values. \Cref{fig:t_schedule} visualizes the change of variables discussed in \cref{sec:parameterization}. All reported results use $c=1$, resulting in uniform schedule, for both training and sampling, except for \noisepred{} and \denoising{} both of which resulted in slightly better FID values for $c=2$ during sampling.

Each model was trained on Google Cloud TPU v3 with $8 \times 8$ configuration. For $2000$ epochs, the $64 \times 64$ model took 2 days to train, $128 \times 128$ took 4 days to train and $256 \times 256$ took 7 days to train. For $1000$ epochs, the training times were roughly the half of that for $2000$ epochs. The training times for the models reported in \cref{tab:resolution_fid} are roughly similar for similarly sized models. Note that our training setup is not maximally optimized for training throughput.

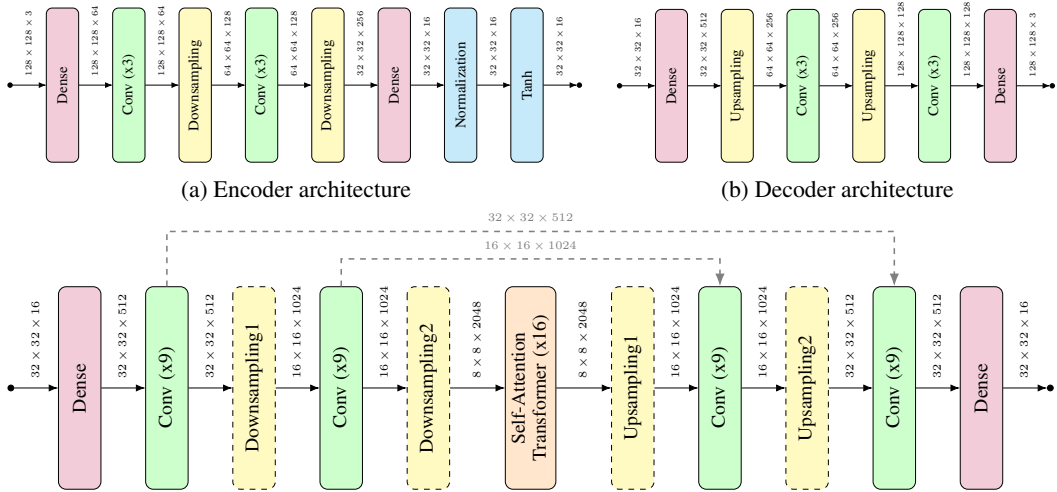
\begin{figure}[t]
\centering
\begin{subfigure}[b]{0.55\linewidth}
\resizebox{\textwidth}{!}{\begin{tikzpicture}[
    auto,
    % Node styles
    io_dot/.style={circle, draw, fill=black, inner sep=0pt, minimum size=3pt},
    % Styles for rotated blocks:
    % Using the same larger dimensions as the final encoder diagram
    block_base/.style={rectangle, draw, minimum width=10.5em, text width=9.75em, minimum height=2.2em, text centered, rounded corners, rotate=90},
    conv_block/.style={block_base, fill=green!20},
    dense_block/.style={block_base, fill=purple!20},
    downsample_block/.style={block_base, fill=yellow!30}, 
    norm_activation_block/.style={block_base, fill=cyan!20}, % New style for InstanceNorm and Tanh
    % Line styles
    line/.style={draw, -Latex},
    % Label styles
    arrow_label/.style={font=\tiny, auto=false, rotate=90, anchor=south}
]

% Define nodes - Horizontal layout
\node[io_dot] (input_dot) {}; 

% Input Dense Block
\node[dense_block, right=0.8cm of input_dot.east, anchor=north] (dense_in) {Dense}; 

% Part 1
\node[conv_block, right=0.8cm of dense_in.south, anchor=north] (conv1_part1) {Conv (x3)}; 
\node[downsample_block, right=0.8cm of conv1_part1.south, anchor=north] (downsample1) {Downsampling};

% Part 2
\node[conv_block, right=0.8cm of downsample1.south, anchor=north] (conv1_part2) {Conv (x3)}; 
\node[downsample_block, right=0.8cm of conv1_part2.south, anchor=north] (downsample2) {Downsampling};

% Output Dense Block
\node[dense_block, right=0.8cm of downsample2.south, anchor=north] (dense_out) {Dense};

% New InstanceNorm Block
\node[norm_activation_block, right=0.8cm of dense_out.south, anchor=north] (instancenorm) {Normalization};

% New Tanh Block
\node[norm_activation_block, right=0.8cm of instancenorm.south, anchor=north] (tanh) {Tanh};

% Output Dot
\node[io_dot, right=0.8cm of tanh.south, anchor=west] (output_dot) {}; 

% Draw main connections with feature map sizes (labels are vertical)
% Anchors adjusted for rotated blocks
\path [line] (input_dot.east) -- node[arrow_label, midway, right, xshift=1mm] {$128 \times 128 \times 3$} (dense_in.north);
\path [line] (dense_in.south) -- node[arrow_label, midway, right, xshift=1mm] {$128 \times 128 \times 64$} (conv1_part1.north);
\path [line] (conv1_part1.south) -- node[arrow_label, midway, right, xshift=1mm] {$128 \times 128 \times 64$} (downsample1.north);
\path [line] (downsample1.south) -- node[arrow_label, midway, right, xshift=1mm] {$64 \times 64 \times 128$} (conv1_part2.north);
\path [line] (conv1_part2.south) -- node[arrow_label, midway, right, xshift=1mm] {$64 \times 64 \times 128$} (downsample2.north);
\path [line] (downsample2.south) -- node[arrow_label, midway, right, xshift=1mm] {$32 \times 32 \times 256$} (dense_out.north);
\path [line] (dense_out.south) -- node[arrow_label, midway, right, xshift=1mm] {$32 \times 32 \times 16$} (instancenorm.north);
\path [line] (instancenorm.south) -- node[arrow_label, midway, right, xshift=1mm] {$32 \times 32 \times 16$} (tanh.north);
\path [line] (tanh.south) -- node[arrow_label, midway, right, xshift=1mm] {$32 \times 32 \times 16$} (output_dot.west);

\end{tikzpicture}}
\caption{Encoder architecture}
\label{fig:encoder_arch}
\end{subfigure}
\hfill
\begin{subfigure}[b]{0.42\linewidth}
\resizebox{\textwidth}{!}{\begin{tikzpicture}[
    auto,
    % Node styles
    io_dot/.style={circle, draw, fill=black, inner sep=0pt, minimum size=3pt},
    % Styles for rotated blocks:
    block_base/.style={rectangle, draw, minimum width=10.5em, text width=9.75em, minimum height=2.2em, text centered, rounded corners, rotate=90},
    conv_block/.style={block_base, fill=green!20},
    dense_block/.style={block_base, fill=purple!20},
    upsample_block/.style={block_base, fill=yellow!30},
    % Line styles
    line/.style={draw, -Latex},
    % Label styles
    arrow_label/.style={font=\tiny, auto=false, rotate=90, anchor=south}
]

\node[io_dot] (input_dot_dec) {}; 

\node[dense_block, right=0.8cm of input_dot_dec.east, anchor=north] (dense_in_dec) {Dense}; 

\node[upsample_block, right=0.8cm of dense_in_dec.south, anchor=north] (upsample1_dec) {Upsampling};
\node[conv_block, right=0.8cm of upsample1_dec.south, anchor=north] (conv1_part1_dec) {Conv (x3)}; 

\node[upsample_block, right=0.8cm of conv1_part1_dec.south, anchor=north] (upsample2_dec) {Upsampling};
\node[conv_block, right=0.8cm of upsample2_dec.south, anchor=north] (conv1_part2_dec) {Conv (x3)}; 

\node[dense_block, right=0.8cm of conv1_part2_dec.south, anchor=north] (dense_out_dec) {Dense};

\node[io_dot, right=0.8cm of dense_out_dec.south, anchor=west] (output_dot_dec) {}; 

\path [line] (input_dot_dec.east) -- node[arrow_label, midway, right, xshift=1mm] {$32 \times 32 \times 16$} (dense_in_dec.north);
\path [line] (dense_in_dec.south) -- node[arrow_label, midway, right, xshift=1mm] {$32 \times 32 \times 512$} (upsample1_dec.north);
\path [line] (upsample1_dec.south) -- node[arrow_label, midway, right, xshift=1mm] {$64 \times 64 \times 256$} (conv1_part1_dec.north);
\path [line] (conv1_part1_dec.south) -- node[arrow_label, midway, right, xshift=1mm] {$64 \times 64 \times 256$} (upsample2_dec.north);
\path [line] (upsample2_dec.south) -- node[arrow_label, midway, right, xshift=1mm] {$128 \times 128 \times 128$} (conv1_part2_dec.north);
\path [line] (conv1_part2_dec.south) -- node[arrow_label, midway, right, xshift=1mm] {$128 \times 128 \times 128$} (dense_out_dec.north);
\path [line] (dense_out_dec.south) -- node[arrow_label, midway, right, xshift=1mm] {$128 \times 128 \times 3$} (output_dot_dec.west);

\end{tikzpicture}}
\caption{Decoder architecture}
\label{fig:decoder_arch}
\end{subfigure}
\newline
\begin{subfigure}[b]{\linewidth}
\resizebox{\textwidth}{!}{\begin{tikzpicture}[
    auto,
    % Node styles
    io_dot/.style={circle, draw, fill=black, inner sep=0pt, minimum size=3pt},
    % Styles for rotated blocks:
    block_base/.style={rectangle, draw, minimum width=10.5em, text width=9.75em, minimum height=2.2em, text centered, rounded corners, rotate=90},
    conv_block/.style={block_base, fill=green!20},
    attn_block/.style={block_base, fill=orange!20}, 
    dense_block/.style={block_base, fill=purple!20},
    up_down_block/.style={block_base, fill=yellow!30}, % This style is now effectively replaced by optional_block_style for these instances
    optional_block_style/.style={block_base, fill=yellow!30, draw=black, dashed}, % For optional blocks
    % Line styles
    line/.style={draw, -Latex},
    shortcut/.style={draw, -Latex, dashed, gray, thick},
    % Label styles
    arrow_label/.style={font=\tiny, auto=false, rotate=90, anchor=south}, 
    shortcut_data_label/.style={font=\tiny, midway, above, yshift=0.25mm} 
]

% Define nodes - Horizontal layout
\node[io_dot] (input) {}; 

\node[dense_block, right=0.8cm of input.east, anchor=north] (dense_in) {Dense}; 

% --- Encoder Part ---
\node[conv_block, right=0.8cm of dense_in.south, anchor=north] (conv_group1_down) {Conv (x9)}; 
% Renamed and numbered
\node[optional_block_style, right=0.8cm of conv_group1_down.south, anchor=north] (downsample1) {Downsampling1};
\node[conv_block, right=0.8cm of downsample1.south, anchor=north] (conv_group2_down) {Conv (x9)}; 

% --- Pre-Attention Downsampling ---
% Renamed and numbered
\node[optional_block_style, right=0.8cm of conv_group2_down.south, anchor=north] (downsample2) {Downsampling2};

% --- Bottleneck ---
\node[attn_block, right=1cm of downsample2.south, anchor=north] (attention) {Self-Attention Transformer (x16)}; 

% --- Post-Attention Upsampling ---
% Renamed and numbered
\node[optional_block_style, right=1cm of attention.south, anchor=north] (upsample1) {Upsampling1};

% --- Decoder Part ---
\node[conv_block, right=0.8cm of upsample1.south, anchor=north] (conv_group1_up) {Conv (x9)}; 
% Renamed and numbered
\node[optional_block_style, right=0.8cm of conv_group1_up.south, anchor=north] (upsample2) {Upsampling2};
\node[conv_block, right=0.8cm of upsample2.south, anchor=north] (conv_group2_up) {Conv (x9)}; 

\node[dense_block, right=0.8cm of conv_group2_up.south, anchor=north] (dense_out) {Dense};
\node[io_dot, right=0.8cm of dense_out.south, anchor=west] (output) {}; 

% Draw main connections with feature map sizes
\path [line] (input.east) -- node[arrow_label, midway, right, xshift=0.5mm] {$32 \times 32 \times 16$} (dense_in.north);
\path [line] (dense_in.south) -- node[arrow_label, midway, right, xshift=0.5mm] {$32 \times 32 \times 512$} (conv_group1_down.north);
\path [line] (conv_group1_down.south) -- node[arrow_label, midway, right, xshift=0.5mm] {$32 \times 32 \times 512$} (downsample1.north);
\path [line] (downsample1.south) -- node[arrow_label, midway, right, xshift=0.5mm] {$16 \times 16 \times 1024$} (conv_group2_down.north);
\path [line] (conv_group2_down.south) -- node[arrow_label, midway, right, xshift=0.5mm] {$16 \times 16 \times 1024$} (downsample2.north);
\path [line] (downsample2.south) -- node[arrow_label, midway, right, xshift=0.5mm] {$8 \times 8 \times 2048$} (attention.north); % Assuming factor 2 downsample & channel doubling
\path [line] (attention.south) -- node[arrow_label, midway, right, xshift=0.5mm] {$8 \times 8 \times 2048$} (upsample1.north);
\path [line] (upsample1.south) -- node[arrow_label, midway, right, xshift=0.5mm] {$16 \times 16 \times 1024$} (conv_group1_up.north); % Upsampled back
\path [line] (conv_group1_up.south) -- node[arrow_label, midway, right, xshift=0.5mm] {$16 \times 16 \times 1024$} (upsample2.north);
\path [line] (upsample2.south) -- node[arrow_label, midway, right, xshift=0.5mm] {$32 \times 32 \times 512$} (conv_group2_up.north);
\path [line] (conv_group2_up.south) -- node[arrow_label, midway, right, xshift=0.5mm] {$32 \times 32 \times 512$} (dense_out.north);
\path [line] (dense_out.south) -- node[arrow_label, midway, right, xshift=0.5mm] {$32 \times 32 \times 16$} (output.west);

% Define vertical offsets for shortcuts (to go above the main flow)
\newdimen\shortcuttwooffsety
\shortcuttwooffsety=0.5cm 

\newdimen\shortcutoneoffsety
\shortcutoneoffsety=1.0cm 

% Draw shortcut connections (labels remain horizontal using shortcut_data_label)
% Shortcut paths remain between the original corresponding convolutional blocks
\coordinate (s2_source_anchor) at (conv_group2_down.east); 
\coordinate (s2_target_anchor) at (conv_group1_up.east);   
\coordinate (s2_p1) at ($(s2_source_anchor) + (0, \shortcuttwooffsety)$); 
\coordinate (s2_p2) at ($(s2_target_anchor)!(s2_p1)!(s2_target_anchor)$); 
\draw [shortcut] (s2_source_anchor) -- (s2_p1) 
    -- node[shortcut_data_label] {$16 \times 16 \times 1024$} (s2_p2) 
    -- (s2_target_anchor); 

\coordinate (s1_source_anchor) at (conv_group1_down.east); 
\coordinate (s1_target_anchor) at (conv_group2_up.east);   
\coordinate (s1_p1) at ($(s1_source_anchor) + (0, \shortcutoneoffsety)$); 
\coordinate (s1_p2) at ($(s1_target_anchor)!(s1_p1)!(s1_target_anchor)$); 
\draw [shortcut] (s1_source_anchor) -- (s1_p1) 
    -- node[shortcut_data_label] {$32 \times 32 \times 512$} (s1_p2) 
    -- (s1_target_anchor);

\end{tikzpicture}}
\caption{Latent stochastic interpolant model architecture. The blocks shown with dashed boundaries are optional across different resolutions.}
\label{fig:lsi_arch}
\end{subfigure}
\caption{An overview of the architecture of various components for $128 \times 128$ resolution model. The architecture for $64 \times 64$ and $256 \times 256$ resolutions is similar, except for the difference in the spatial feature map sizes. See \cref{sec:arch_details_app} for details.}
\label{fig:architecture_overfiew_app}
\end{figure}

%%%%%%%%%%%%%%%%%%%%%%%%%%%%%%%%%%%%%%%%%%%%%%%%%%%%%%%%%%%%%%%%%%%%%
\section{Architecture details}
\label{sec:arch_details_app}
The base architecture of our model is adapted from the work described by \citet{hoogeboom2023simple} and modified to separate out Encoder, Decoder and Latent SI models. In the adapted base architecture feature maps are processed using groups of convolution blocks and downsampled spatially after each group, to yield the lowest feature map resolution at $16 \times 16$. A sequence of Self-Attention Transformer blocks then operates on the $16 \times 16$ feature map. Note that the transformer blocks in our adapted architecture operate only at $16 \times 16$ resolution. Consequently, for a $64 \times 64$ resolution input image, two downsamplings are performed, for $128 \times 128$ resolution, three downsamplings are performed and for $256 \times 256$ four downsamplings are performed.  All convolutional groups have the same number of convolutional blocks. The observation space SI models used in this paper are constructed using this adapted base architecture. To construct Encoder, Decoder and Latent SI models, we simply partition the base model into three parts. The first part contains two groups of convolutional blocks, each followed by downsampling, and forms the encoder. An extra dense layer is added to reduce the number of channels. Further, the output is normalized to have zero mean and unit standard deviation followed by tanh activation to limit the range to $[-1, 1]$. Similarly, the last part contains two groups of convolutional blocks, each followed by upsampling, and forms the decoder. An extra dense layer is added at the beginning to increase the number of channels. The remaining middle portion forms the Latent SI model, where two extra dense layers are added, one at beginning and one at end to increase and decrease the feature map sizes respectively. We show an overview of the architecture for various components in the \cref{fig:architecture_overfiew_app}.

Note that the \texttt{tanh} activation or other forms of scale control, such as normalization, play a crucial role in preventing the encoder from learning arbitrarily large embeddings and allowing it to achieve better FID. Without this constraint, the model makes the encoder outputs have large scale to make denoising easier at later timesteps. This is an important implementation detail that ensures stable training. Empirically, encoder output normalization yielded more stable training and better FID, than without anything, at the same number of steps. Addition of \texttt{tanh} further improved the FID.

For different resolutions, the Encoder and Decoder models are fully convolutional and have the same architecture. The architecture of Latent SI models differs in the presence/absence of the optional downsampling and upsampling blocks (shown as blocks with dashed boundaries). The $64 \times 64$ Latent SI model does not contain any downsampling/upsampling blocks as the encoder output is already $16 \times 16$. The $128 \times 128$ model does not contain "Downsampling1" and "Upsampling2" blocks. The $256 \times 256$ model contains all blocks. All models contain 16 Self-Attention Transformer blocks. To increase/decrease number of parameters to match model capacities, only the number of convolutional blocks in groups immediately before and after the Self-Attention Transformer blocks is changed.

All models operate with a $3\times$ smaller latent dimensionality that the observations. We focused on this dimensionality ratio to ensure fair comparison with observation-space baselines while maintaining reasonable latent dimensionality for effective modeling. In earlier experiments we tried other compression ratios including $2\times$ and $4\times$, before settling on $3\times$. The primary effect of the dimensionality ratio is on the reconstruction performance. Higher the dimensionality ratio, the harder it is for the decoder to achieve a high PSNR at the same number of training steps, resulting in worse sample quality (FID) and longer training times. Lower the dimensionality ratio, less the computational advantage.

\section{Additional sampling details and results}
\label{sec:sampling_details_app}
\begin{figure}[htbp]
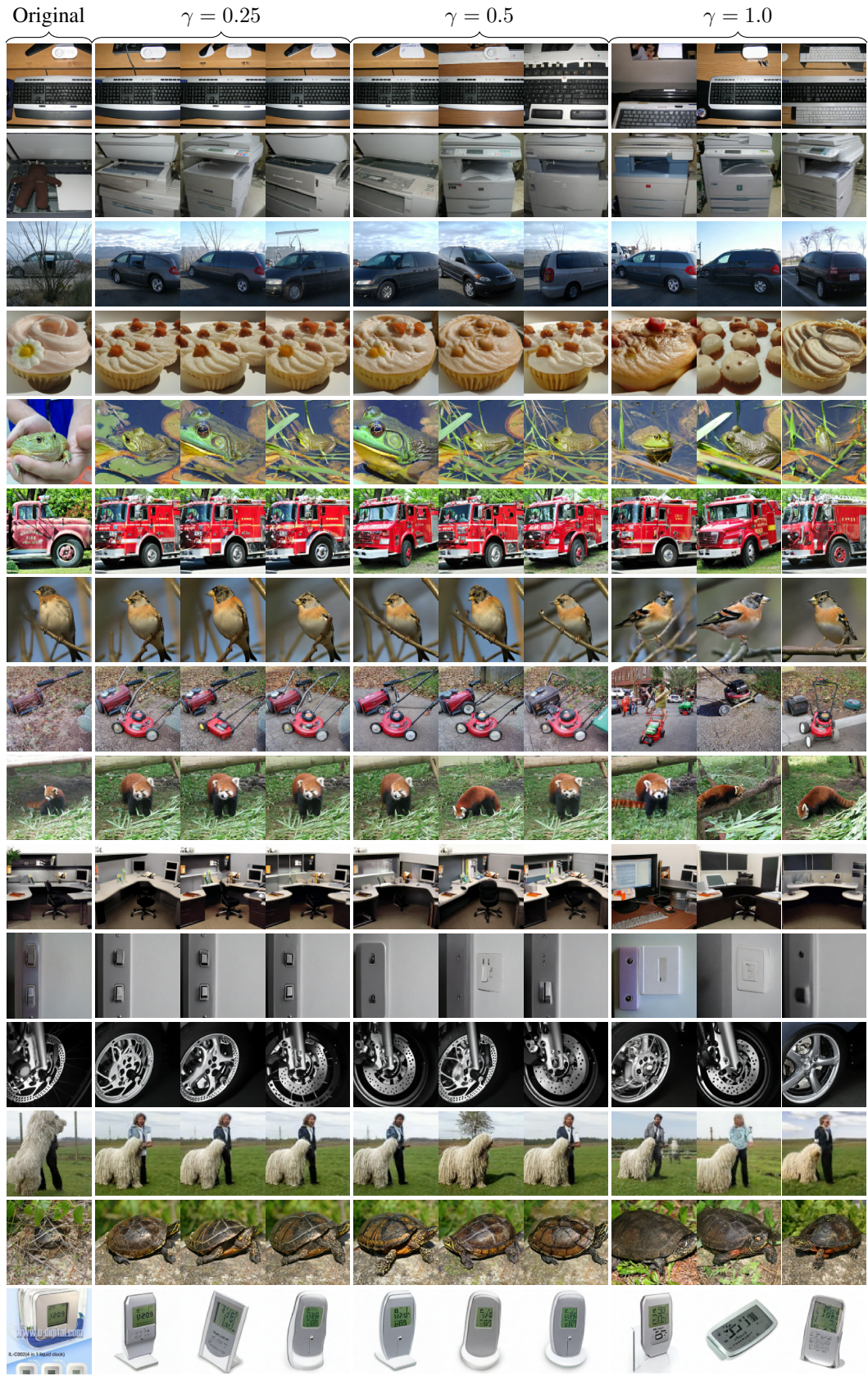
 % htbp are placement specifiers (here, top, bottom, page)
    \centering % Center the figure content

    \begin{tikzpicture}[
        imgnode/.style={inner sep=0pt, outer sep=0pt}, % Style for image nodes to remove padding
        brace_style/.style={decorate, decoration={brace, amplitude=5pt}} % Style for upward braces, reduced amplitude
      ]
        % Node for the first PDF (first row) - Index 6
        \node[imgnode] (img1) {\includegraphics[width=\textwidth, page=1]{\imagepath gamma_grid_0.pdf}};

        % --- Labels ---
        % Labels are positioned with their baseline (anchor=south) 4pt above img1.north.
        % This accounts for a 3pt brace amplitude + 1pt gap.
        \newcommand{\labelYShift}{4pt} % 3pt brace amplitude + 1pt gap

        \node[anchor=south, yshift=\labelYShift] at ($(img1.north west)!0.05!(img1.north east)$) {Original};      % Centered over the first 10% (1 image)
        \node[anchor=south, yshift=\labelYShift] at ($(img1.north west)!0.25!(img1.north east)$) {$\gamma=0.25$}; % Centered over 10%-40% (3 images)
        \node[anchor=south, yshift=\labelYShift] at ($(img1.north west)!0.55!(img1.north east)$) {$\gamma=0.5$};  % Centered over 40%-70% (3 images)
        \node[anchor=south, yshift=\labelYShift] at ($(img1.north west)!0.85!(img1.north east)$) {$\gamma=1.0$};  % Centered over 70%-100% (3 images)

        % --- Braces ---
        % Braces are drawn directly along the top edge of the image (img1.north), pointing upwards.
        % The points for the brace path are on img1.north.
        \draw [brace_style]
            ($(img1.north west)$) -- ($(img1.north west)!0.1!(img1.north east)$);

        \draw [brace_style]
            ($(img1.north west)!0.1!(img1.north east)$) -- ($(img1.north west)!0.4!(img1.north east)$);

        \draw [brace_style]
            ($(img1.north west)!0.4!(img1.north east)$) -- ($(img1.north west)!0.7!(img1.north east)$);

        \draw [brace_style]
            ($(img1.north west)!0.7!(img1.north east)$) -- ($(img1.north east)$);

        % --- Subsequent Images ---
        % Nodes for the subsequent PDFs, positioned with uniform spacing.
        % Using anchor=north and below=<distance> of <previous_node>.south for consistent visual gaps.
        \newcommand{\rowspacing}{0.1em} % Define uniform row spacing - REDUCED for closer rows

        \node[imgnode, anchor=north, below=\rowspacing of img1.south] (img2) {\includegraphics[width=\textwidth, page=1]{\imagepath gamma_grid_9.pdf}};
        
        \node[imgnode, anchor=north, below=\rowspacing of img2.south] (img3) {\includegraphics[width=\textwidth, page=1]{\imagepath gamma_grid_10.pdf}};
        
        \node[imgnode, anchor=north, below=\rowspacing of img3.south] (img4) {\includegraphics[width=\textwidth, page=1]{\imagepath gamma_grid_13.pdf}};
        
        \node[imgnode, anchor=north, below=\rowspacing of img4.south] (img5) {\includegraphics[width=\textwidth, page=1]{\imagepath gamma_grid_15.pdf}};

        \node[imgnode, anchor=north, below=\rowspacing of img5.south] (img6) {\includegraphics[width=\textwidth, page=1]{\imagepath gamma_grid_23.pdf}};

        \node[imgnode, anchor=north, below=\rowspacing of img6.south] (img7) {\includegraphics[width=\textwidth, page=1]{\imagepath gamma_grid_24.pdf}};

        \node[imgnode, anchor=north, below=\rowspacing of img7.south] (img8) {\includegraphics[width=\textwidth, page=1]{\imagepath gamma_grid_25.pdf}};

        \node[imgnode, anchor=north, below=\rowspacing of img8.south] (img9) {\includegraphics[width=\textwidth, page=1]{\imagepath gamma_grid_27.pdf}};

        \node[imgnode, anchor=north, below=\rowspacing of img9.south] (img10) {\includegraphics[width=\textwidth, page=1]{\imagepath gamma_grid_30.pdf}};

        \node[imgnode, anchor=north, below=\rowspacing of img10.south] (img11) {\includegraphics[width=\textwidth, page=1]{\imagepath gamma_grid_33.pdf}};

        \node[imgnode, anchor=north, below=\rowspacing of img11.south] (img12) {\includegraphics[width=\textwidth, page=1]{\imagepath gamma_grid_38.pdf}};

        \node[imgnode, anchor=north, below=\rowspacing of img12.south] (img13) {\includegraphics[width=\textwidth, page=1]{\imagepath gamma_grid_45.pdf}};

        \node[imgnode, anchor=north, below=\rowspacing of img13.south] (img14) {\includegraphics[width=\textwidth, page=1]{\imagepath gamma_grid_60.pdf}};

        \node[imgnode, anchor=north, below=\rowspacing of img14.south] (img15) {\includegraphics[width=\textwidth, page=1]{\imagepath gamma_grid_74.pdf}};

    \end{tikzpicture}

    \caption{LSI supports flexible sampling.}
    \label{fig:gamma_grid_app}
\end{figure}

\begin{figure}[htbp]
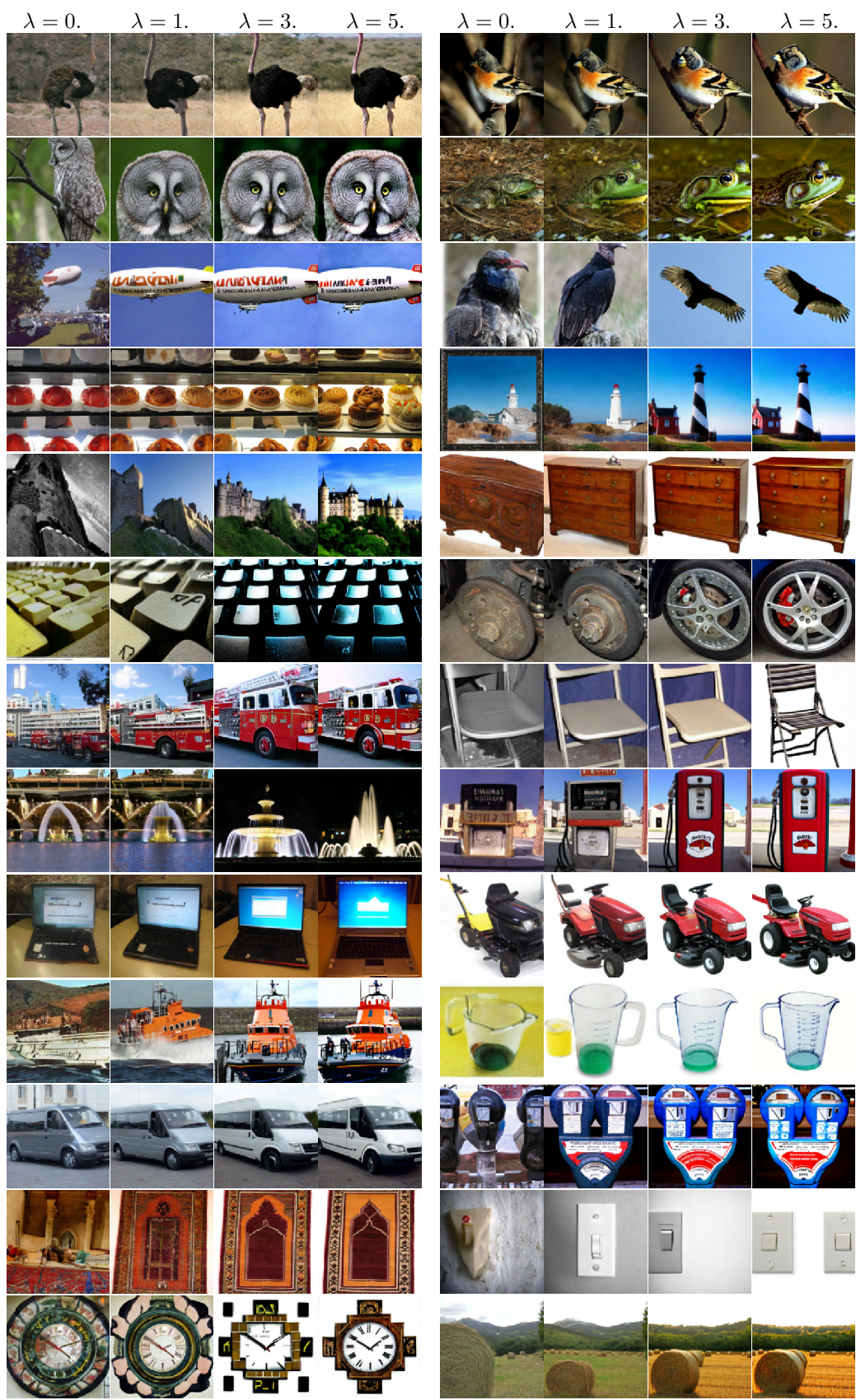
 % 'h'ere, 't'op, 'b'ottom, 'p'age of floats
    \centering
    \begin{tikzpicture}
        % Define the width of each image (adjust as needed)
        \def\imagewidth{0.48\textwidth}

        % --- Coordinates for Spacing ---
        \def\newxdist{7.0} % Horizontal distance for the second column
        \def\firstrowy{1.5} % Y-coordinate for the center of the first row images
        \def\rowspacing{1.7} % Vertical distance between the centers of rows (adjust if needed)

        \node (img11) at (0, \firstrowy) {\includegraphics[width=\imagewidth]{\imagepath cfg_9_122.pdf}};
        \node (img12) at (\newxdist, \firstrowy) {\includegraphics[width=\imagewidth]{\imagepath cfg_10_24.pdf}};

        \node (img21) at (0, \firstrowy - \rowspacing) {\includegraphics[width=\imagewidth]{\imagepath cfg_24_21.pdf}};
        \node (img22) at (\newxdist, \firstrowy - \rowspacing) {\includegraphics[width=\imagewidth]{\imagepath cfg_30_15.pdf}};

        \node (img31) at (0, \firstrowy - 2*\rowspacing) {\includegraphics[width=\imagewidth]{\imagepath cfg_405_32.pdf}};
        \node (img32) at (\newxdist, \firstrowy - 2*\rowspacing) {\includegraphics[width=\imagewidth]{\imagepath cfg_23_94.pdf}};

        \node (img41) at (0, \firstrowy - 3*\rowspacing) {\includegraphics[width=\imagewidth]{\imagepath cfg_415_13.pdf}};
        \node (img42) at (\newxdist, \firstrowy - 3*\rowspacing) {\includegraphics[width=\imagewidth]{\imagepath cfg_437_105.pdf}};

        \node (img51) at (0, \firstrowy - 4*\rowspacing) {\includegraphics[width=\imagewidth]{\imagepath cfg_483_6.pdf}};
        \node (img52) at (\newxdist, \firstrowy - 4*\rowspacing) {\includegraphics[width=\imagewidth]{\imagepath cfg_493_46.pdf}};

        \node (img61) at (0, \firstrowy - 5*\rowspacing) {\includegraphics[width=\imagewidth]{\imagepath cfg_508_0.pdf}};
        \node (img62) at (\newxdist, \firstrowy - 5*\rowspacing) {\includegraphics[width=\imagewidth]{\imagepath cfg_535_38.pdf}};

        \node (img71) at (0, \firstrowy - 6*\rowspacing) {\includegraphics[width=\imagewidth]{\imagepath cfg_555_23.pdf}};
        \node (img72) at (\newxdist, \firstrowy - 6*\rowspacing) {\includegraphics[width=\imagewidth]{\imagepath cfg_559_56.pdf}};

        \node (img81) at (0, \firstrowy - 7*\rowspacing) {\includegraphics[width=\imagewidth]{\imagepath cfg_562_19.pdf}};
        \node (img82) at (\newxdist, \firstrowy - 7*\rowspacing) {\includegraphics[width=\imagewidth]{\imagepath cfg_571_100.pdf}};

        \node (img91) at (0, \firstrowy - 8*\rowspacing) {\includegraphics[width=\imagewidth]{\imagepath cfg_620_85.pdf}};
        \node (img92) at (\newxdist, \firstrowy - 8*\rowspacing) {\includegraphics[width=\imagewidth]{\imagepath cfg_621_25.pdf}};

        \node (img101) at (0, \firstrowy - 9*\rowspacing) {\includegraphics[width=\imagewidth]{\imagepath cfg_625_59.pdf}};
        \node (img102) at (\newxdist, \firstrowy - 9*\rowspacing) {\includegraphics[width=\imagewidth]{\imagepath cfg_647_64.pdf}};

        \node (img111) at (0, \firstrowy - 10*\rowspacing) {\includegraphics[width=\imagewidth]{\imagepath cfg_654_112.pdf}};
        \node (img112) at (\newxdist, \firstrowy - 10*\rowspacing) {\includegraphics[width=\imagewidth]{\imagepath cfg_704_14.pdf}};

        \node (img121) at (0, \firstrowy - 11*\rowspacing) {\includegraphics[width=\imagewidth]{\imagepath cfg_741_103.pdf}};
        \node (img122) at (\newxdist, \firstrowy - 11*\rowspacing) {\includegraphics[width=\imagewidth]{\imagepath cfg_844_33.pdf}};

        \node (img131) at (0, \firstrowy - 12*\rowspacing) {\includegraphics[width=\imagewidth]{\imagepath cfg_892_17.pdf}};
        \node (img132) at (\newxdist, \firstrowy - 12*\rowspacing) {\includegraphics[width=\imagewidth]{\imagepath cfg_958_51.pdf}};

        % --- Add Labels Above Sub-Images in the First Row ---
        % (This part remains the same as before)
        \coordinate (img11_sub1_top) at ($(img11.north west)!0.125!(img11.north east)$);
        \coordinate (img11_sub2_top) at ($(img11.north west)!0.375!(img11.north east)$);
        \coordinate (img11_sub3_top) at ($(img11.north west)!0.625!(img11.north east)$);
        \coordinate (img11_sub4_top) at ($(img11.north west)!0.875!(img11.north east)$);

        \node[above=-5pt of img11_sub1_top] {$\lambda=0.$}; % Replace 1.1
        \node[above=-5pt of img11_sub2_top] {$\lambda=1.$}; % Replace 1.2
        \node[above=-5pt of img11_sub3_top] {$\lambda=3.$}; % Replace 1.3
        \node[above=-5pt of img11_sub4_top] {$\lambda=5.$}; % Replace 1.4

        \coordinate (img12_sub1_top) at ($(img12.north west)!0.125!(img12.north east)$);
        \coordinate (img12_sub2_top) at ($(img12.north west)!0.375!(img12.north east)$);
        \coordinate (img12_sub3_top) at ($(img12.north west)!0.625!(img12.north east)$);
        \coordinate (img12_sub4_top) at ($(img12.north west)!0.875!(img12.north east)$);

        \node[above=-5pt of img12_sub1_top] {$\lambda=0.$}; % Replace 2.1
        \node[above=-5pt of img12_sub2_top] {$\lambda=1.$}; % Replace 2.2
        \node[above=-5pt of img12_sub3_top] {$\lambda=3.$}; % Replace 2.3
        \node[above=-5pt of img12_sub4_top] {$\lambda=5.$}; % Replace 2.4

    \end{tikzpicture}
    \caption{LSI supports CFG sampling.}
    \label{fig:cfg_sampling_app}
\end{figure}

All the results reported in the paper use the deterministic sampler with 300 steps, setting $\gamma_t = 0$ in \cref{eq:general_sampler}, except when otherwise stated. \cref{fig:gamma_grid} and \cref{fig:gamma_grid_app} use stochastic sampling with $\gamma_t \equiv \gamma (1-t)$, where $\gamma$ is a specified constant. We use Euler (for probability flow ODE) and Euler-Maruyama (for SDE) discretization for all results, except for qualitative inversion results in \cref{fig:gamma_grid} and \cref{fig:gamma_grid_app}. For the inversion results we experimented with two reversible samplers: 1) Reversible Heun \citep{kidger2021efficient} and, 2) Asynchronous Leapfrog Integrator \citep{zhuang2021mali}. While both exhibited instability and failed to invert some of the images, we found  Asynchronous Leapfrog Integrator to be more stable in our experiments and used it for results in \cref{fig:gamma_grid} and \cref{fig:gamma_grid_app}. \Cref{fig:cfg_sampling_app} provides additional samples for qualitative assessment, complementing \cref{fig:cfg_sampling} in the main paper.

Sampling speed (with 100 steps) for pixel space models is roughly 2.2 images/sec/core for 64x64, 0.95 images/sec/core for 128x128 and 0.21 images/sec/core for 256x256. LSI achieves 2.65 images/sec/core for 64x64, 1.30 images/sec/core, and 0.53 images/sec/core for 256x256. We would like to emphasize that these numbers exhibit high variance, are highly hardware dependent and can be significantly impacted by hardware specific optimizations that are not the focus of this paper.

\begin{table}[t]
\centering
\caption{Comparison with state-of-the-art FID results on ImageNet 128$\times$128. Note that these
models have differing sizes, FLOPs and NFEs. The comparison is provided purely for
reference.}
\label{tab:sota_comparison_app}
\begin{tabular}{l c}
\toprule
Method & FID \\
\midrule
Ours & 3.12 \\
\midrule
SiD2 \citep{hoogeboom2024simpler} & 1.26 \\
PaGoDA \citep{kim2024pagoda} & 1.48 \\
DisCo-Diff \citep{xu2024disco} & 1.73 \\
VDM++ \citep{kingma2023understanding} & 1.75  \\
SiD \citep{hoogeboom2023simple} & 1.94 \\ 
RIN \citep{jabri2022scalable} & 2.75 \\
CDM \citep{ho2022classifier} & 3.52 \\
ADM \citep{dhariwal2021diffusion} & 5.91 \\
\bottomrule
\end{tabular}%
\end{table}

%%%%%%%%%%%%%%%%%%%%%%%%%%%%%%%%%%%%%%%%%%%%%%%%%%%%%%%%%%%%%%%%%%%%%
\section{Comparison with other methods}
\label{sec:other_comparison_app}

While the primary focus of this paper is on the theoretical results and their empirical
validation, in \cref{tab:sota_comparison_app} we present comparison with other image generation methods for completeness. We provide this table purely for reference as these methods are not directly comparable due to differing model sizes, FLOPs and NFEs. While our best result
is comparable, techniques in these works are complementary to our method. We leave it
as future work to explore this direction.

%%%%%%%%%%%%%%%%%%%%%%%%%%%%%%%%%%%%%%%%%%%%%%%%%%%%%%%%%%%%%%%%%%%%%
\section{Use of LLM}
\label{sec:use_of_llm_app}
LLMs were used to help create some of the figures in the paper.

\end{document}